\documentclass[letterpaper]{article} % DO NOT CHANGE THIS
\usepackage[draft]{aaai2026}  % DO NOT CHANGE THIS
\usepackage{times}  % DO NOT CHANGE THIS
\usepackage{helvet}  % DO NOT CHANGE THIS
\usepackage{courier}  % DO NOT CHANGE THIS
\usepackage[hyphens]{url}  % DO NOT CHANGE THIS
\usepackage{graphicx} % DO NOT CHANGE THIS
\usepackage{natbib}  % DO NOT CHANGE THIS AND DO NOT ADD ANY OPTIONS TO IT
\usepackage{caption} % DO NOT CHANGE THIS AND DO NOT ADD ANY OPTIONS TO IT
\usepackage{algorithm}
\usepackage{algorithmic}

\usepackage{newfloat}
\usepackage{listings}
\DeclareCaptionStyle{ruled}{labelfont=normalfont,labelsep=colon,strut=off} % DO NOT CHANGE THIS
\floatstyle{ruled}
\newfloat{listing}{tb}{lst}{}
\floatname{listing}{Listing}
\usepackage{amsmath}
\usepackage[most]{tcolorbox}

\usepackage{longtable}
\usepackage{color}
\usepackage{booktabs}
\usepackage{multirow}
\usepackage{pifont}
\usepackage{hyperref}

\usepackage{appendix}
\title{Are All Prompt Components Value-Neutral? Understanding the Heterogeneous Adversarial Robustness of Dissected Prompt in Large Language Models}
\author {
    Yujia Zheng\textsuperscript{\rm 1}\thanks{Equal contribution, co-first author.},
    Tianhao Li\textsuperscript{\rm 1}\footnotemark[1], \\
    Haotian Huang\textsuperscript{\rm 2}, 
    Tianyu Zeng\textsuperscript{\rm 3}, 
    Jingyu Lu\textsuperscript{\rm 4}, 
    Chuangxin Chu\textsuperscript{\rm 5}, \\
    Yuekai Huang\textsuperscript{\rm 6,7}, 
    Ziyou Jiang\textsuperscript{\rm 6,7}, 
    Qian Xiong\textsuperscript{\rm 8}, 
    Yuyao Ge\textsuperscript{\rm 7,9}, 
    Mingyang Li\textsuperscript{\rm 6,7}\thanks{Corresponding author.}
}
\affiliations {
    \textsuperscript{\rm 1}Duke University, 
    \textsuperscript{\rm 2}North China University of Technology
    \textsuperscript{\rm 3}Hong Kong Polytechnic University
    \textsuperscript{\rm 4}Australian National University
    \textsuperscript{\rm 5}Nanyang Technological University
    \textsuperscript{\rm 6}Institute of Software, Chinese Academy of Sciences
    \textsuperscript{\rm 7}University of Chinese Academy of Sciences
    \textsuperscript{\rm 8}Beijing Forestry University
    \textsuperscript{\rm 9}Institute of Computing Technology, Chinese Academy of Sciences
    \\
    \texttt{\{yujia.zheng, tianhao.li\}@duke.edu,  mingyang2017@iscas.ac.cn}
}
\usepackage{bibentry}
\begin{document}

\maketitle

\begin{abstract}
\vspace{-0.5em}
Prompt-based adversarial attacks have become an effective means to assess the robustness of large language models (LLMs). However, existing approaches often treat prompts as monolithic text, overlooking their structural heterogeneity—different prompt components contribute unequally to adversarial robustness. Prior works like PromptRobust assume prompts are value-neutral, but our analysis reveals that complex, domain-specific prompts with rich structures have components with differing vulnerabilities. To address this gap, we introduce \textsc{PromptAnatomy}, an automated framework that dissects prompts into functional components and generates diverse, interpretable adversarial examples by selectively perturbing each component using our proposed method, \textsc{ComPerturb}. To ensure linguistic plausibility and mitigate distribution shifts, we further incorporate a perplexity (PPL)-based filtering mechanism. As a complementary resource, we annotate four public instruction-tuning datasets using the \textsc{PromptAnatomy} framework, verified through human review. Extensive experiments across these datasets and five advanced LLMs demonstrate that \textsc{ComPerturb} achieves state-of-the-art attack success rates. Ablation studies validate the complementary benefits of prompt dissection and PPL filtering. Our results underscore the importance of prompt structure awareness and controlled perturbation for reliable adversarial robustness evaluation in LLMs. Code and data are available at \href{https://github.com/Yujiaaaaa/PACP}{https://github.com/Yujiaaaaa/PACP}. 
\end{abstract}

% Uncomment the following to link to your code, datasets, an extended version or similar.
% You must keep this block between (not within) the abstract and the main body of the paper.
% \begin{links}
%     \link{Code}{https://aaai.org/example/code}
%     \link{Datasets}{https://aaai.org/example/datasets}
%     \link{Extended version}{https://aaai.org/example/extended-version}
% \end{links}

\vspace{-1.5em}
\section{Introduction}

Large language models (LLMs) \citep{naveed2023comprehensive, zhao2023survey} such as DeepSeek, ChatGPT, and LLaMA-3 have demonstrated remarkable capabilities in a wide range of tasks \citep{thirunavukarasu2023large, zhang2025scientific, demszky2023using, hou2024large}, largely driven by in-context learning via carefully crafted prompts \citep{dong2022survey, li2023practical, long2024prompt, mei2025survey}. As these models become increasingly integrated into real-world applications, the robustness of their behavior under perturbed prompts is becoming a concern for both reliability and safety \citep{zhu2023promptbench, zhu2023promptrobust, hu2024prompt, ghosh2025ailuminate}.

Recent studies have demonstrated that even subtle variations in prompt phrasing—such as typos, synonym substitutions, or formatting changes—can lead to significant drops in LLM performance, a phenomenon known as prompt sensitivity \citep{zhu2023promptbench, zhu2023promptrobust, wang2023mttm}. To evaluate this vulnerability, \citet{zhu2023promptrobust} introduced \textit{PromptRobust}, a benchmark targeting adversarial prompt perturbations, while \citet{wang2023mttm} proposed \textit{MTTM}, a metamorphic testing framework that applies structured perturbations to toxic inputs for testing content moderation systems. Although these efforts highlight the fragility of LLMs under adversarial prompting, they treat prompts as flat, undifferentiated text—\textbf{overlooking the functional structure within prompts that may contribute unequally to model behavior}.

\begin{figure}[t]
    \centering
    \includegraphics[width=\linewidth]{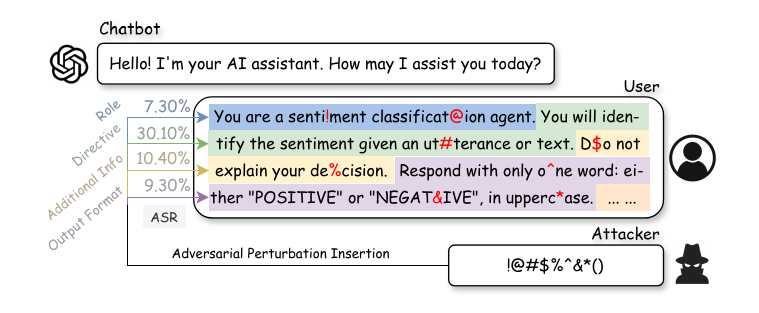}
    \vspace{-1.5em}
    \caption{Inserting adversarial perturbations into different prompt components leads to distinct attack success rates (ASR).}
    \label{fig:pilot}
    \vspace{-1.5em}
\end{figure}

We argue that \textbf{prompts are fundamentally compositional, consisting of multiple functional components}—such as task directives, role assignments, input-output delimiters, exemplars, and answer formatting instructions—each guiding LLMs to interpret and execute tasks more effectively. In practical scenarios, users often construct complex prompts by layering basic instructions with supplementary cues, such as specifying the task domain, assigning a role to the model (e.g., “You are a medical expert”), providing concrete examples, or defining output styles. Recognizing the diversity and importance of these components, \citet{schulhoff2024prompt} provide a comprehensive taxonomy that categorizes prompt elements into six types: \textit{Directive}, \textit{Examples}, \textit{Output Formatting}, \textit{Style Instructions}, \textit{Role}, and \textit{Additional Information}. To further streamline the taxonomy for our analysis, we merge \textit{Style Instructions} into \textit{Output Formatting}, as both primarily define the expected form or tone of the model’s output. These components exhibit heterogeneous adversarial robustness, meaning that some are more vulnerable to perturbations than others. To investigate this, we pose the following research questions:
{\textit{RQ1}}: How does prompt structure and component richness affect the effectiveness of adversarial perturbations?
{\textit{RQ2}}: How does the sensitivity of adversarial robustness vary across different prompt component types?
{\textit{RQ3}}: How do semantic and syntactic perturbations differ in effectiveness across prompt components?

To address these questions, we introduce \textsc{PromptAnatomy}, \textbf{the first framework for structurally decomposing prompts into canonical components}. Our framework achieves superior accuracy compared to GPT-4o \citep{hurst2024gpt} in dissecting long, complex prompt structures.  This dissection enables fine-grained analysis of how each component contributes to model robustness. We first complete missing components and then apply \textsc{PromptAnatomy} to four complex, domain-specific prompt datasets, resulting in structurally annotated variants (\textit{PubMedQA-PA, EMEA-PA, Leetcode-PA, CodeGeneration-PA}). Each dataset undergoes manual review by human experts, achieving 100\% inter-annotator agreement to ensure annotation quality.

Building on this foundation, we propose \textsc{ComPerturb}, \textbf{a component-wise adversarial perturbation method} that targets each prompt component with customized perturbation strategies. This enables controlled robustness evaluation by isolating the effects of specific prompt modifications. We conduct extensive experiments across multiple LLMs and tasks, revealing that certain components are significantly more vulnerable to perturbation than others in specific circumstances. Our findings provide actionable insights for prompt engineers and model developers to craft safer, more reliable prompts in high-stakes applications.

\paragraph{Contributions.} This paper makes the following key contributions:
% \begin{itemize}
    (\romannumeral1) We introduce \textsc{PromptAnatomy}, the first framework for structurally decomposing prompts for LLMs into multiple canonical components, enabling fine-grained analysis and systematic prompt design.
    (\romannumeral2) We apply \textsc{PromptAnatomy} to four complex domain-specific prompt datasets, producing variants annotated with structural components (denoted with the suffix '-PA'). Each dataset is manually reviewed to ensure annotation quality, achieving 100\% agreement among human annotators.
    (\romannumeral3) We introduce \textsc{ComPerturb}, a component-wise perturbation method that applies targeted adversarial strategies to individual prompt components, enabling fine-grained analysis of their impact on model behavior and adversarial robustness.
    (\romannumeral4) We conduct comprehensive experiments across diverse LLMs and datasets based on \textsc{ComPerturb}, demonstrating which components most significantly influence robustness. Our findings offer actionable insights for prompt engineers and model developers to design safer and more reliable prompting strategies.
% \end{itemize}

\paragraph{Findings.} We summarize three key findings from our comprehensive evaluation:
(\romannumeral1) \textbf{Component and Structure-Guided Perturbations Amplify Adversarial Effectiveness in Complex Prompts.} \textsc{ComPerturb} performs best on structurally rich prompts, showing that component-aware attacks are especially powerful in complex settings. (\textit{RQ1})
(\romannumeral2) \textbf{Prompt Components Exhibit Heterogeneous Robustness to Adversarial Perturbations.} Components like \textit{DIR} and \textit{ADI} are more susceptible to perturbations, while \textit{ROL} and \textit{OFT} remain relatively robust. (\textit{RQ2})
(\romannumeral3) \textbf{Semantic Perturbations Are More Effective Than Syntactic Ones Across Prompt Components.} Meaning-level changes outperform syntactic ones across components, highlighting the role of perturbation type. (\textit{RQ3}) Base on these findings, We provide guidelines and takeaways for four potential target audience: offensive/defensive security researchers, prompt engineers, developers, and general users.

\section{Related Works}

\paragraph{Prompt Dissection}
Prompt dissection breaks prompts into functional components—such as directives, roles, and formatting—that shape LLM behavior \citep{schulhoff2024prompt}. This modular approach improves control, interpretability, and task-solving by leveraging reusable substructures \citep{khot2022decomposed}, and reveals latent multilingual capabilities in LLMs \citep{nie2024decomposed}. In adversarial contexts, it enhances attack effectiveness through targeted reconstruction \citep{li2024drattack}, underscoring the value of component-aware prompt analysis.

\paragraph{Adversarial Robustness}
Prompt robustness addresses the resilience of LLMs to adversarial manipulations and variations in prompt wording and structure \citep{wang2023robustness}. Due to the sensitivity of LLM outputs to minor perturbations in prompts, ensuring robustness has become a critical area of research, particularly in safety-sensitive domains such as healthcare, finance, and science. Additionally, recent advancements extend robustness considerations to LLM-based agents, reflecting the increasing complexity and interactive contexts of deployed systems \citep{xiong2025butterfly, hu2025enhancing}. everal works propose black-box adversarial attacks that exploit prompt-level vulnerabilities \citep{zhu2023promptbench, zou2023universal, xue2023trojllm, das2024human}, while others investigate certified robustness or ensemble-based defenses \citep{aguilera2025llm}. These studies underscore the importance of evaluating LLMs under structurally-aware perturbation paradigms beyond token-level noise.

\paragraph{LLM Safety Benchmark} 

% A range of benchmarks assess LLM safety and robustness. SecReEvalBench targets multi-turn security vulnerabilities \citep{cui2025secreevalbench}, while SG-Bench evaluates safety generalization across prompt formats \citep{mou2024sg}. ChemSafetyBench addresses hazardous chemistry jailbreaks \citep{zhao2024chemsafetybench}, and R‑Judge tests risk detection in agent interactions \citep{yuan2024r}. Agent-SafetyBench spans 349 environments and 10 failure modes \citep{zhang2024agent}, and SALAD-Bench offers hierarchical safety evaluations across NLU tasks \citep{li2024salad}. SciSafeEval focuses on safety alignment in scientific domains \citep{li2024scisafeeval}. Other benchmarks include AdvGLUE for adversarial robustness \citep{wang2021adversarial} and MASK for probing honesty under pressure \citep{ren2025mask}.

A range of safety benchmarks has been developed to evaluate LLM alignment across adversarial, ethical, and domain-specific risks \citep{lu2025alignment}. General benchmarks like AdvBench \citep{wang2021adversarial}, SafetyBench \citep{zhang2023safetybench}, and SALAD-Bench \citep{li2024salad} assess refusal behaviors and robustness to jailbreaks. Domain-specific benchmarks—ChemSafetyBench \citep{zhao2024chemsafetybench}, MedSafetyBench \citep{han2024medsafetybench}, and SciSafeEval \citep{li2024scisafeeval}—test models 
 on scientific misuse, ethical compliance, and jailbreak resistance. Agent-based evaluations like Agent-SafetyBench \citep{zhang2024agent} and SafeAgentBench \citep{yin2024safeagentbench} measure risks in dynamic tool-use environments. While these benchmarks standardize vulnerability assessment, they generally overlook the compositional structure of prompts.

\section{Methodology}

\subsection{\textsc{PromptAnatomy}: Identifying and Dissecting LLM Prompt Components}

\begin{figure*}[t]
    \centering
    \includegraphics[width=\linewidth]{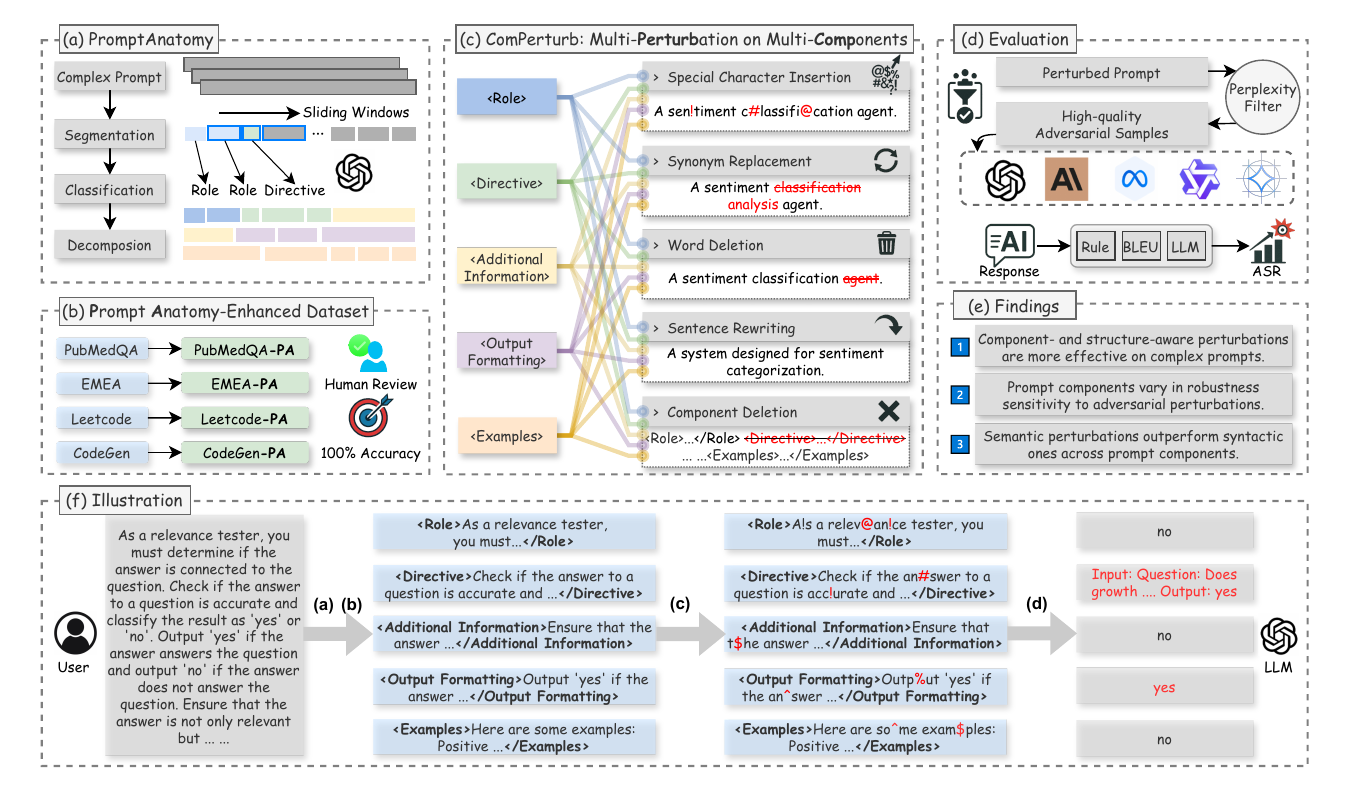}
    \caption{Overview of \textsc{PromptAnatomy} and \textsc{ComPerturb}. 
        (a) Origin prompts are dissected into components using a sliding window LLM classifier.
        (b) Dissection is applied to benchmarks to create annotated datasets. 
        (c) \textsc{ComPerturb} applies perturbations to components.
        (d) Perturbed prompts are enhanced and use to evaluate LLMs.
        (e) Analysis reveals component-wise vulnerabilities. 
        (f) A working example shows dissection, perturbation, and output shifts.
        }
    \label{fig:prompt_anatomy}
\end{figure*}

\paragraph{Formulation} 
Let $\mathcal{P}$ denote the space of free-form natural language prompts. Each prompt $p \in \mathcal{P}$ is typically an unstructured token sequence designed to elicit LLM behavior. Existing methods treat $p$ as a monolithic input, lacking a formal structure for semantic analysis. We define \textit{prompt dissection} as mapping $p$ into a structured sequence of functional components:
\[
\mathcal{A}: \mathcal{P} \rightarrow \mathcal{C}^n,
\]
where $\mathcal{C}$ is the set of canonical components and $n$ is the number of identified components. The output $\mathcal{A}(p) = (c_1, c_2, \dots, c_n)$ represents the decomposed prompt under our framework, \textsc{PromptAnatomy}, which enables component-aware analysis and robustness evaluation.

\paragraph{Challenges}
\textit{Prompt dissection} introduces several practical and technical challenges. First, existing human-in-the-loop methods~\citep{shivagunde2024deconstructing} rely heavily on manual annotation, which is time-consuming and infeasible at scale. Second, LLMs often exhibit limited robustness when parsing long or complex prompts. For instance, they may omit content, oversimplify components, or focus disproportionately on the beginning and end of the prompt, neglecting critical middle sections (see examples in the Appendix \ref{sec:llm_fails_to_decompose}). Third, current LLMs struggle with disentangling overlapping or implicitly expressed components, leading to incorrect or incomplete dissection. This issue is amplified when the interpretation of a sentence depends on adjacent content (e.g., “Use the following format”), highlighting the need for context-aware classification. Finally, the absence of publicly available benchmark datasets tailored for prompt dissection impedes both method development and standardized evaluation.

\paragraph{Limited Context-Aware Prompt Dissection}

To address these challenges, we develop a two-stage dissection pipeline (Algorithm~\ref{alg:prompt_decompose}) that combines rule-based segmentation with LLM-based classification under contextual guidance. First, each raw prompt is segmented into individual sentences using punctuation delimiters (e.g., periods, question marks) to enable fine-grained analysis. For each sentence, we construct a local context window by collecting up to two preceding and two following sentences, which helps disambiguate meaning in context-dependent cases. Each target sentence and its context are then passed to a language model (e.g., GPT-4o) using a structured prompt that instructs the model to classify the sentence into one or more of five canonical components: \texttt{<Role>}, \texttt{<Directive>}, \texttt{<Additional Information>}, \texttt{<Output Formatting>}, and \texttt{<Examples>}. The model’s responses are aggregated and post-processed to ensure consistency, with an additional verification step to identify and reclassify any missing sentences. The final output is a structured representation of the prompt, where each sentence is wrapped in XML-style tags corresponding to its identified component(s).

\begin{algorithm}[htbp]
\caption{Prompt Dissection Pipeline}
\label{alg:prompt_decompose}
\begin{algorithmic}[1]
\REQUIRE Prompt $p$ from dataset $\mathcal{D}$
\ENSURE Component tags $\mathcal{A}(p)$

\STATE $S \gets \text{Split}(p, \texttt{[".", "?"]})$
\STATE $\mathcal{A}(p) \gets \emptyset$

\FOR{$i = 1$ to $|S|$}
    \STATE $C_i \gets \text{ContextWindow}(S, i, k=2)$
    \STATE $r_i \gets \text{LLMClassify}(s_i, C_i)$
    \STATE $\mathcal{A}(p) \gets \mathcal{A}(p) \cup \{r_i\}$
\ENDFOR

\STATE $S_{\text{miss}} \gets \text{FindUnlabeled}(S, \mathcal{A}(p))$

\FOR{$s_j \in S_{\text{miss}}$}
    \STATE $C_j \gets \text{ContextWindow}(S, j, k=2)$
    \STATE $r_j \gets \text{LLMClassify}(s_j, C_j)$
    \STATE $\mathcal{A}(p) \gets \mathcal{A}(p) \cup \{r_j\}$
\ENDFOR

\RETURN $\mathcal{A}(p)$
\end{algorithmic}
\end{algorithm}

As shown in Table~\ref{tab:classification_accuracy} in the Appendix \ref{sec:acc}, \textsc{PromptAnatomy} shows strong reliability, achieving 100\% accuracy on GPT-4o and over 90\% on other open-weight models across diverse domains. We applied it to a diverse set of existing prompt-based datasets across multiple domains, including biomedical question answering (\textit{PubMedQA} \citep{wang2022super}), translation (\textit{EMEA} \citep{wang2022super}), calculating password strength (\textit{Leetcode} \citep{wang2022super}), and code generation (\textit{CodeGeneration} \citep{nceyda2025yap470}). For each dataset, we automatically dissect the raw prompts into structured components, which result in enhanced datasets \textit{PubMedQA-PA}, \textit{EMEA-PA}, \textit{Leetcode-PA}, \textit{and CodeGeneration-PA} (see Appendix \ref{sec:datasets} for details).

\subsection{\textsc{ComPerturb}: Component-wise Perturbation Method on LLM Prompt}

Grounded in our \textsc{PromptAnatomy} dissection framework, we propose a novel \textbf{com}ponent-wise \textbf{perturb}ation method \textsc{ComPerturb}. Formally, given a prompt $p$ dissected into a set of $k$ labeled components $\{c_1, c_2, \dots, c_k\}$, \textsc{CompPerturb} defines an adversarial transformation function $\mathcal{T}_i$ for each component $c_i$, resulting in a perturbed prompt $p^{(i)} = \{c_1, \dots, \mathcal{T}_i(c_i), \dots, c_k\}$. Each $\mathcal{T}_i$ is instantiated with a perturbation strategy tailored to the semantics of component $c_i$. We then evaluate the robustness of an LLM $f$ under each component-wise perturbation by measuring the change in output $\Delta f(p, p^{(i)})$ using task-specific metrics. 

To instantiate $\mathcal{T}_i$ for each component $c_i$, we design a suite of multi-level perturbation operations that introduce varying degrees of semantic distortion: (\romannumeral1) special character insertion, which injects noise symbols (e.g., ‘@’, ‘\#’, ‘!’) into words to simulate typos or obfuscation; (\romannumeral2) synonym replacement, which substitutes words with semantically similar alternatives using \texttt{nltk}; (\romannumeral3) word deletion, which removes selected tokens to mimic underspecified or incomplete prompts; (\romannumeral4) sentence rewriting, which replaces a full sentence with a paraphrased variant to simulate natural reformulations; and (\romannumeral5) component deletion, which removes an entire functional segment of the prompt (e.g., \texttt{<Directive>} or \texttt{<Examples>}) based on our structural decomposition.

To ensure the effectiveness and representativeness of adversarial examples, we apply a perplexity-guided filtering strategy. Specifically, for each perturbed prompt $p^{(i)}$, we compute its perplexity $\text{PPL}(p^{(i)})$ using a reference language model. To quantify the relative semantic disturbance, we calculate the {perplexity ratio} between the perturbed prompt and its corresponding clean version $p^{(0)}$ as:

\begin{align}
\text{Ratio}^{(i)} = \frac{\text{PPL}\left(p^{(i)}\right)}{\text{PPL}\left(p^{(0)}\right)}
\end{align}

where $\operatorname{Ratio}^{(i)}$ denotes the perplexity ratio of the $i$-th perturbed prompt, $p^{(i)}$ is the perturbed version of the prompt, and $p^{(0)}$ is the original unperturbed prompt.

We then rank all perturbed samples in descending order of $\operatorname{Ratio}^{(i)}$ and retain the top 20\% as our final high-quality adversarial set, effectively filtering out low-impact or semantically trivial perturbations.

\section{Experiment}

\paragraph{Setup.} We evaluate the effectiveness of \textsc{CompPerturb} across a diverse suite of language models and task domains. For target models, we select five widely-used LLMs with varying capacities and architectures: \textit{GPT-4o} (M1), \textit{Claude3.7} (M2), \textit{Qwen2.5-14B} (M3), \textit{LLaMA3.3-70B} (M4), and \textit{Gemma3-12B} (M5). For adversarial prompt generation, we employ both commercial and open-weight models, including \textit{GPT-4o}, \textit{GPT-3.5}, \textit{Qwen3-8B}, \textit{Qwen2.5-14B}, \textit{LLaMA3.3-70B}, and \textit{Gemma3-12B}. Our evaluation uses datasets from distinct domains: \textit{PubMedQA-PA} (PM-PA), \textit{EMEA-PA} (EM-PA), \textit{Leetcode-PA} (LC-PA), and \textit{CodeGeneration-PA} (CG-PA). Inference was performed using 4 NVIDIA RTX™ A6000 GPUs for open-weight models, and through official APIs for commercial models. We compare our method against baselines \textit{MTTM}~\citep{wang2023mttm} and \textit{PromptRobust}~\citep{zhu2023promptrobust}.

\begin{table*}[t]
\centering
\small
\setlength{\tabcolsep}{4pt}
\renewcommand{\arraystretch}{1}
{\fontsize{8pt}{9pt}\selectfont
\begin{tabular}{lcccccc}
\toprule
\textbf{Method} & \textbf{GPT-4o} & \textbf{Claude3.7} & \textbf{Qwen2.5-14B} & \textbf{LLaMA3.3-70B} & \textbf{Gemma3-12B} & \textbf{Average} \\
\midrule
\multicolumn{7}{c}{\textbf{PubMedQA-PA} (PCM = 34.883; AVG $\uparrow$=17.76\%)} \\
\midrule
MTTM & 42.50 (40.70$\pm$0.02) & 46.88 (44.23$\pm$0.02) & 44.67 (39.33$\pm$0.03) & 45.74 (41.74$\pm$0.02) & 47.32 (39.08$\pm$0.01) & 45.42 (41.02$\pm$0.02) \\
PromptRobust & 44.37 (41.56$\pm$0.03) & 47.34 (45.34$\pm$0.07) & 46.75 (41.71$\pm$0.02) & 49.03 (47.03$\pm$0.05) & 49.75 (46.25$\pm$0.03) & 47.45 (44.38$\pm$0.04) \\
ComPerturb (Ours) & 64.98 (53.63$\pm$0.07) & 67.01 (60.02$\pm$0.04) & 60.94 (50.76$\pm$0.06) & 65.04 (52.38$\pm$0.07) & 63.00 (51.73$\pm$0.07) & 64.19 (53.70$\pm$0.06) \\
\midrule
\multicolumn{7}{c}{\textbf{EMEA-PA} (PCM = 32.100; AVG $\uparrow$=10.24\%)} \\
\midrule
MTTM & 62.30 (60.01$\pm$0.02) & 62.24 (60.26$\pm$0.01) & 74.20 (69.97$\pm$0.02) & 73.52 (72.00$\pm$0.01) & 76.30 (72.89$\pm$0.02) & 69.72 (67.03$\pm$0.02) \\
PromptRobust & 69.30 (58.74$\pm$0.04) & 67.50 (61.56$\pm$0.03) & 72.30 (70.34$\pm$0.05) & 75.70 (72.87$\pm$0.06) & 74.60 (73.81$\pm$0.03) & 71.88 (67.46$\pm$0.04) \\
ComPerturb (Ours) & 75.90 (67.10$\pm$0.05) & 78.00 (69.11$\pm$0.05) & 84.90 (78.09$\pm$0.04) & 82.12 (75.93$\pm$0.04) & 84.30 (78.66$\pm$0.03) & 81.04 (73.78$\pm$0.04) \\
\midrule
\multicolumn{7}{c}{\textbf{Leetcode-PA} (PCM = 41.495; AVG $\uparrow$=20.16\%)} \\
\midrule
MTTM & 57.20 (53.45$\pm$0.02) & 52.30 (49.10$\pm$0.01) & 47.90 (47.96$\pm$0.03) & 54.70 (51.89$\pm$0.03) & 54.40 (50.88$\pm$0.02) & 53.30 (50.66$\pm$0.02) \\
PromptRobust & 56.30 (54.98$\pm$0.05) & 55.20 (54.32$\pm$0.02) & 56.70 (55.12$\pm$0.05) & 51.20 (50.32$\pm$0.03) & 56.90 (54.30$\pm$0.02) & 55.26 (53.81$\pm$0.03) \\
ComPerturb (Ours) & 81.50 (75.40$\pm$0.04) & 72.08 (65.74$\pm$0.04) & 65.30 (62.57$\pm$0.02) & 78.80 (76.10$\pm$0.03) & 74.50 (67.42$\pm$0.04) & 74.44 (69.45$\pm$0.03) \\
\midrule
\multicolumn{7}{c}{\textbf{CodeGeneration-PA} (PCM = 44.524; AVG $\uparrow$=29.6\%)} \\
\midrule
MTTM & 31.20 (27.98$\pm$0.02) & 29.30 (27.77$\pm$0.01) & 40.20 (38.40$\pm$0.03) & 40.30 (38.17$\pm$0.03) & 41.70 (38.78$\pm$0.02) & 36.54 (34.22$\pm$0.02) \\
PromptRobust & 29.08 (23.46$\pm$0.05) & 29.30 (27.32$\pm$0.02) & 40.20 (35.78$\pm$0.05) & 41.20 (36.21$\pm$0.03) & 38.40 (33.21$\pm$0.02) & 35.78 (31.20$\pm$0.03) \\
ComPerturb (Ours) & 66.96 (61.48$\pm$0.04) & 64.00 (57.09$\pm$0.04) & 66.36 (59.35$\pm$0.05) & 65.40 (60.72$\pm$0.03) & 66.08 (60.19$\pm$0.03) & 65.76 (59.77$\pm$0.04) \\
\bottomrule
\end{tabular}
}
\caption{Comparison of ASR (\%) of \textsc{ComPerturb} and baselines across five LLMs and four datasets. Each cell reports the best ASR, followed by the mean ASR and standard deviation from different perturbation strategies of each method.}
\label{tab:merged_baseline}
\end{table*}

\paragraph{Metrics.} 
We adopt task-specific metrics. For classification-style tasks such as PubMedQA and Leetcode, we define an attack as successful if the LLM's output no longer contains the correct answer. For generation tasks such as translation, we calculate Bilingual Evaluation Understudy (BLEU) scores \citep{papineni2002bleu} to measure the semantic preservation or degradation after perturbation, an attack is considered a success if BLEU is less than 20, which indicate poor translation quality \citep{koehn2017six}. For code generation tasks (evaluated on the Code dataset), we employ GPT-4o as an automatic reference model to assess the correctness of the generated code. 

Based on various successful metrics for different tasks, we compute Attack Success Rate (ASR) as:

\begin{align}
\text{ASR} = \frac{N_{\text{success}}}{N_{\text{total}}} \times 100\%
\end{align}

where $N_{success}$ is the number of samples attacked successfully, $N_{total}$ is the total number of samples.

We introduce the \textbf{Prompt Complexity Metric (PCM)}, a quantitative measure designed to evaluate the inherent complexity of LLM prompts. PCM integrates five core dimensions: lexical rarity, syntactic depth, semantic dispersion, structural richness, and task difficulty. Each component captures a distinct aspect of prompt formulation. The PCM score is computed as a weighted sum of these dimensions:

\begin{align}
\text{PCM}(p) =\; & \alpha C_{\text{lexical}} + \beta C_{\text{syntactic}} + \gamma C_{\text{semantic}} \notag \\\
                 & + \delta C_{\text{structural}} + \epsilon C_{\text{task}},
\end{align}

where $C_{\text{lexical}}$ denotes the average inverse document frequency of tokens, $C_{\text{syntactic}}$ reflects syntactic tree depth, $C_{\text{semantic}}$ measures pairwise embedding-based semantic distance, $C_{\text{structural}}$ captures the organization of prompt components, and $C_{\text{task}}$ encodes the inherent difficulty of the task. Detail could be found at Appendix \ref{sec:pcm}.

\section{Results and Analysis}

Table~\ref{tab:merged_baseline} provides a comparison of proposed \textsc{ComPerturb} with baselines. For each method, we report both the highest ASR and the mean ASR with standard deviation, aggregated over multiple perturbation strategies. The results demonstrate that \textsc{ComPerturb} achieves the highest ASRs across all settings. It achieves significant gains in complex scenarios by leveraging a component-aware design that targets semantically critical prompt segments, highlighting the advantage of structurally and semantically informed perturbations. % Transition here

\begin{tcolorbox}[
  colback=pink!15,
  colframe=black!40,
  boxrule=1pt,
  left=2pt, right=2pt, top=2pt, bottom=2pt,
  sharp corners,
  enhanced,
  drop shadow
]
\textbf{Finding \#1}: Component and Structure-Guided Perturbations Result in a Better Effect in Complex Prompts.
\end{tcolorbox}

Based on this finding, we derive a practical guideline to improve LLM robustness for developers.

\begin{tcolorbox}[
  colback=blue!10!white,
  colframe=black!40,
  boxrule=1pt,
  left=2pt, right=2pt, top=2pt, bottom=2pt,
  sharp corners,
  enhanced,
  drop shadow
]
\textbf{Guideline \#1}: Enhancing robustness requires clear task specification by users and component-aware adversarial data augmentation by developers.
\end{tcolorbox}

\begin{table*}[htbp]
\centering
\small
\setlength{\tabcolsep}{1.1pt}
\renewcommand{\arraystretch}{1}
% \resizebox{\textwidth}{!}{ % <-- This line scales table to fit width
\begin{tabular}{l|lc | *{5}{c} | *{5}{c} | *{5}{c} | *{5}{c}}
\toprule
\textbf{Model}                & \textbf{COM} & \textbf{PPL} & \multicolumn{5}{c|}{\textbf{PubMedQA-PA}} & \multicolumn{5}{c|}{\textbf{EMEA-PA}}                                    & \multicolumn{5}{c|}{\textbf{Leetcode-PA}} & \multicolumn{5}{c}{\textbf{CodeGeneration-PA}} \\ \cmidrule(l){4-23} 
                              &              &              & SCI    & SYR    & WOD    & SER    & COD   & SCI  & SYR                      & WOD                      & SER  & COD  & SCI    & SYR    & WOD    & SER    & COD   & SCI     & SYR     & WOD     & SER     & COD    \\ \midrule
\multirow{11}{*}{GPT-4o}      & ROL          & w/o          & 46.0   & 40.0   & 42.5   & 46.0   & 42.5  & 58.5 & 58.5                     & 57.5                     & 57.5 & 61.0 & 57.4   & 60.0   & 60.5   & 65.4   & 65.0  & 43.5    & 43.5    & 42.5    & 40.5    & 41.5   \\
                              &              & w/           & 47.0   & 45.5   & 52.5   & 50.0   & 47.4  & 65.0 & 59.0                     & 60.0                     & 57.5 & 67.5 & 70.5   & 70.5   & 75.0   & 69.0   & 75.0  & 60.0    & 52.5    & 57.4    & 60.0    & 62.5   \\
                              & DIR          & w/o          & 56.0   & 52.0   & 59.5   & 50.0   & 77.5  & 62.5 & 62.0                     & 72.0                     & 64.5 & 99.5 & 72.5   & 70.0   & 72.0   & 72.5   & 95.0  & 50.5    & 47.0    & 49.5    & 47.5    & 64.0   \\
                              &              & w/           & 60.0   & 57.4   & 64.0   & 65.0   & 78.5  & 67.5 & 67.5                     & 73.0                     & 72.5 & 99.5 & 75.0   & 77.5   & 80.0   & 77.5   & 97.5  & 67.0    & 62.5    & 67.0    & 65.0    & 72.5   \\
                              & ADI          & w/o          & 46.0   & 43.0   & 47.5   & 45.0   & 52.5  & 59.0 & 61.0                     & 59.5                     & 62.5 & 59.5 & 62.5   & 67.5   & 71.0   & 70.0   & 77.5  & 44.0    & 43.5    & 45.0    & 45.0    & 45.5   \\
                              &              & w/           & 55.0   & 45.0   & 57.4   & 43.5   & 75.0  & 62.5 & 63.5                     & 68.5                     & 67.5 & 68.0 & 70.5   & 68.5   & 80.0   & 72.5   & 87.0  & 67.5    & 60.0    & 62.5    & 65.0    & 62.5   \\
                              & OFT          & w/o          & 45.0   & 46.0   & 45.5   & 48.0   & 47.0  & 59.0 & 61.0                     & 58.5                     & 58.0 & 58.0 & 57.4   & 66.0   & 65.5   & 66.0   & 66.5  & 41.5    & 45.5    & 43.0    & 44.0    & 43.5   \\
                              &              & w/           & 47.5   & 50.0   & 46.0   & 50.0   & 52.5  & 62.5 & 75.0                     & 65.0                     & 62.0 & 65.0 & 71.0   & 75.0   & 77.5   & 69.0   & 85.0  & 65.0    & 57.5    & 57.4    & 62.5    & 55.0   \\
                              & EXA          & w/o          & 45.5   & 43.5   & 42.0   & 47.5   & 46.0  & 57.5 & 59.5                     & 58.0                     & 57.5 & 61.5 & 50.0   & 55.0   & 69.0   & 65.0   & 69.5  & 45.5    & 43.5    & 44.0    & 44.0    & 42.0   \\
                              &              & w/           & 47.5   & 44.0   & 55.0   & 55.0   & 47.0  & 62.5 & 62.5                     & 65.0                     & 62.0 & 77.5 & 68.0   & 64.0   & 69.0   & 67.0   & 97.5  & 62.5    & 57.4    & 62.5    & 57.4    & 55.0   \\
                              & CRT          & w/o          & 44.0   & 43.5   & 43.0   & 48.5   & 52.5  & 60.5 & 59.5                     & 59.5                     & 59.5 & 69.5 & 60.0   & 69.0   & 70.5   & 69.5   & 76.0  & 45.0    & 43.5    & 44.0    & 43.0    & 54.0   \\
                              &              & w/           & 55.0   & 45.0   & 60.0   & 50.0   & 55.0  & 62.5 & 62.5                     & 60.0                     & 70.0 & 70.5 & 71.0   & 70.0   & 72.5   & 70.0   & 87.0  & 62.0    & 60.0    & 65.0    & 55.0    & 62.0   \\ \midrule
\multirow{11}{*}{Claude3.7}   & ROL          & w/o          & 50.0   & 48.0   & 48.0   & 52.0   & 52.5  & 57.5 & 58.0                     & 50.0                     & 58.5 & 59.0 & 50.4   & 53.0   & 50.0   & 50.5   & 50.5  & 39.0    & 40.0    & 42.0    & 39.0    & 39.5   \\
                              &              & w/           & 55.0   & 52.5   & 53.5   & 65.5   & 57.5  & 67.5 & 62.5                     & 59.0                     & 70.0 & 65.0 & 57.4   & 53.5   & 72.5   & 55.0   & 75.0  & 60.0    & 57.4    & 55.0    & 60.0    & 50.0   \\
                              & DIR          & w/o          & 58.5   & 56.0   & 57.0   & 59.5   & 77.5  & 67.0 & 61.0                     & 68.5                     & 64.5 & 99.0 & 60.5   & 59.0   & 61.0   & 60.0   & 90.0  & 46.0    & 49.0    & 49.0    & 46.0    & 67.0   \\
                              &              & w/           & 65.0   & 60.0   & 60.0   & 62.5   & 87.5  & 77.5 & 65.0                     & 75.0                     & 72.5 & 100  & 65.0   & 62.5   & 74.0   & 67.4   & 92.0  & 60.0    & 62.5    & 62.5    & 60.0    & 72.5   \\
                              & ADI          & w/o          & 56.5   & 54.0   & 52.0   & 55.0   & 64.0  & 61.5 & 59.0                     & 60.0                     & 59.5 & 59.5 & 53.0   & 60.5   & 57.4   & 53.5   & 76.5  & 43.0    & 44.0    & 42.0    & 43.0    & 49.0   \\
                              &              & w/           & 58.0   & 57.5   & 55.0   & 65.0   & 65.0  & 67.5 & 72.0                     & 69.5                     & 70.0 & 70.0 & 53.5   & 60.0   & 65.0   & 56.0   & 96.5  & 55.0    & 57.0    & 55.5    & 55.0    & 60.0   \\
                              & OFT          & w/o          & 52.0   & 50.5   & 54.0   & 58.5   & 66.5  & 58.5 & 57.0                     & 60.0                     & 59.5 & 58.5 & 50.0   & 50.5   & 51.5   & 55.0   & 51.5  & 42.0    & 40.0    & 42.0    & 42.0    & 40.0   \\
                              &              & w/           & 55.0   & 57.5   & 54.5   & 60.0   & 66.0  & 70.0 & 67.0                     & 60.0                     & 68.5 & 60.0 & 54.5   & 60.0   & 67.5   & 60.0   & 80.0  & 57.5    & 50.0    & 57.4    & 57.5    & 57.5   \\
                              & EXA          & w/o          & 52.0   & 52.0   & 56.0   & 52.5   & 57.4  & 60.0 & 59.5                     & 59.0                     & 58.0 & 59.0 & 53.0   & 54.5   & 55.5   & 56.4   & 52.5  & 40.0    & 45.0    & 43.0    & 40.0    & 39.0   \\
                              &              & w/           & 55.0   & 60.0   & 57.5   & 57.5   & 58.0  & 65.0 & 65.0                     & 67.5                     & 68.5 & 72.5 & 57.5   & 55.7   & 56.5   & 57.0   & 90.0  & 55.0    & 57.4    & 55.0    & 55.0    & 47.5   \\
                              & CRT          & w/o          & 55.5   & 52.5   & 52.0   & 52.5   & 59.0  & 59.0 & 60.0                     & 60.0                     & 59.0 & 70.5 & 57.4   & 57.4   & 56.9   & 55.0   & 60.5  & 40.0    & 41.0    & 42.0    & 40.0    & 49.0   \\
                              &              & w/           & 57.0   & 55.0   & 52.5   & 55.0   & 70.0  & 67.0 & 70.0                     & 67.5                     & 67.0 & 70.5 & 62.5   & 61.0   & 59.0   & 61.0   & 87.5  & 50.0    & 57.5    & 55.5    & 50.0    & 55.0   \\ \midrule
\multirow{11}{*}{Qwen2.5-14B} & ROL          & w/o          & 39.2   & 44.2   & 42.0   & 46.0   & 38.7  & 71.1 & 74.5                     & 72.5                     & 75.7 & 73.2 & 51.6   & 50.3   & 50.2   & 50.5   & 51.7  & 44.0    & 43.0    & 42.5    & 39.5    & 43.0   \\
                              &              & w/           & 43.7   & 43.7   & 45.0   & 48.5   & 47.5  & 74.5 & 76.5                     & 76.2                     & 77.5 & 75.0 & 49.5   & 53.5   & 62.5   & 64.0   & 70.5  & 60.5    & 57.4    & 57.0    & 57.4    & 60.5   \\
                              & DIR          & w/o          & 41.5   & 50.0   & 47.0   & 47.5   & 96.5  & 74.8 & 77.3                     & 87.0                     & 81.2 & 99.5 & 56.8   & 54.2   & 55.7   & 53.5   & 93.0  & 59.5    & 49.5    & 49.5    & 49.5    & 67.0   \\
                              &              & w/           & 50.0   & 51.2   & 55.0   & 51.2   & 97.5  & 75.0 & 80.0                     & 88.7                     & 81.3 & 99.5 & 60.5   & 54.5   & 64.0   & 54.5   & 93.0  & 70.0    & 57.4    & 68.0    & 59.5    & 77.4   \\
                              & ADI          & w/o          & 41.4   & 47.5   & 45.2   & 46.0   & 40.7  & 74.7 & 75.2                     & 75.0                     & 75.5 & 74.7 & 51.9   & 54.2   & 54.6   & 54.3   & 92.1  & 49.5    & 45.0    & 45.0    & 41.0    & 45.5   \\
                              &              & w/           & 50.0   & 52.0   & 51.5   & 51.2   & 47.5  & 74.8 & 73.8                     & 80.0                     & 72.5 & 89.0 & 52.0   & 56.4   & 57.9   & 55.6   & 98.0  & 62.5    & 50.0    & 62.8    & 56.0    & 62.5   \\
                              & OFT          & w/o          & 42.0   & 45.0   & 43.7   & 44.5   & 45.0  & 74.2 & 74.7                     & 75.2                     & 74.5 & 75.0 & 50.4   & 50.2   & 51.6   & 50.6   & 52.0  & 40.5    & 43.0    & 44.5    & 40.8    & 44.0   \\
                              &              & w/           & 52.5   & 52.5   & 50.0   & 47.5   & 48.7  & 72.5 & 76.3                     & 80.0                     & 75.0 & 80.0 & 45.5   & 51.5   & 74.5   & 52.0   & 85.0  & 72.5    & 52.5    & 57.4    & 58.0    & 53.0   \\
                              & EXA          & w/o          & 41.0   & 45.7   & 45.0   & 45.5   & 37.2  & 71.2 & 76.2                     & 75.0                     & 73.0 & 74.2 & 50.7   & 50.1   & 52.3   & 51.5   & 54.6  & 43.5    & 42.0    & 44.0    & 40.5    & 43.0   \\
                              &              & w/           & 38.7   & 47.5   & 50.0   & 48.5   & 47.5  & 75.0 & 80.0                     & 75.2                     & 74.0 & 70.5 & 50.9   & 51.5   & 62.5   & 56.0   & 88.5  & 57.4    & 55.0    & 60.0    & 50.0    & 50.0   \\
                              & CRT          & w/o          & 39.5   & 47.0   & 43.0   & 45.4   & 47.0  & 73.8 & 75.7                     & 75.7                     & 74.7 & 79.0 & 51.7   & 53.7   & 51.5   & 52.5   & 59.6  & 45.5    & 42.5    & 44.5    & 40.5    & 49.5   \\
                              &              & w/           & 41.2   & 43.7   & 43.5   & 47.0   & 50.0  & 75.0 & 76.0                     & 75.7                     & 76.3 & 78.8 & 56.0   & 53.7   & 62.0   & 54.0   & 90.0  & 60.5    & 62.5    & 60.0    & 52.7    & 60.0   \\ \midrule
\multirow{11}{*}{LLaMA3-70B}  & ROL          & w/o          & 45.0   & 40.0   & 43.5   & 44.0   & 45.0  & 70.5 & 70.0                     & 70.7                     & 72.5 & 71.2 & 70.0   & 60.0   & 66.0   & 66.5   & 78.5  & 40.0    & 42.0    & 45.0    & 41.0    & 43.0   \\
                              &              & w/           & 48.0   & 48.5   & 46.5   & 47.5   & 52.5  & 70.0 & 70.0                     & 71.7                     & 73.7 & 78.7 & 70.0   & 69.5   & 75.0   & 66.0   & 80.0  & 57.4    & 55.0    & 58.0    & 60.0    & 60.0   \\
                              & DIR          & w/o          & 49.0   & 49.5   & 50.0   & 49.5   & 100   & 76.0 & 74.0                     & 83.2                     & 75.5 & 99.7 & 70.0   & 73.0   & 72.5   & 68.5   & 93.5  & 51.5    & 51.0    & 50.0    & 52.5    & 68.5   \\
                              &              & w/           & 57.5   & 52.5   & 58.0   & 57.5   & 100   & 78.7 & 76.0                     & 83.5                     & 73.7 & 99.7 & 75.0   & 72.0   & 75.0   & 78.5   & 93.5  & 62.5    & 62.5    & 65.0    & 62.0    & 75.0   \\
                              & ADI          & w/o          & 47.0   & 45.5   & 46.5   & 47.0   & 47.0  & 72.0 & 71.0                     & 72.2                     & 71.2 & 70.0 & 72.5   & 71.5   & 72.0   & 72.5   & 80.0  & 42.2    & 46.0    & 43.0    & 47.0    & 53.0   \\
                              &              & w/           & 50.0   & 47.5   & 55.0   & 50.0   & 50.0  & 75.0 & 71.3                     & 73.2                     & 75.0 & 82.5 & 75.0   & 77.5   & 72.5   & 72.5   & 90.0  & 70.0    & 62.5    & 62.5    & 52.5    & 60.5   \\
                              & OFT          & w/o          & 48.5   & 45.0   & 43.0   & 46.5   & 46.5  & 72.0 & 71.5                     & 71.2                     & 70.2 & 71.5 & 68.5   & 71.0   & 71.0   & 64.5   & 71.0  & 42.0    & 42.0    & 45.0    & 41.0    & 44.0   \\
                              &              & w/           & 48.5   & 47.0   & 49.5   & 50.0   & 50.0  & 71.4 & 77.8                     & 72.5                     & 75.0 & 75.0 & 72.5   & 75.0   & 87.5   & 65.0   & 80.5  & 65.0    & 57.4    & 60.0    & 62.0    & 57.4   \\
                              & EXA          & w/o          & 46.5   & 44.0   & 46.5   & 45.5   & 38.0  & 70.5 & 70.2                     & 72.0                     & 69.2 & 73.2 & 71.5   & 70.0   & 67.0   & 66.5   & 66.5  & 40.5    & 43.5    & 44.5    & 46.0    & 40.0   \\
                              &              & w/           & 47.0   & 45.0   & 57.4   & 52.0   & 42.5  & 71.5 & 76.5                     & 70.0                     & 73.7 & 85.0 & 72.5   & 70.0   & 80.0   & 65.0   & 92.5  & 57.4    & 62.5    & 62.5    & 56.0    & 52.5   \\
                              & CRT          & w/o          & 47.5   & 47.5   & 43.4   & 46.0   & 54.0  & 73.7 & 71.0                     & 72.0                     & 70.7 & 77.8 & 69.5   & 66.5   & 66.5   & 68.0   & 84.0  & 41.0    & 42.5    & 45.0    & 41.5    & 50.0   \\
                              &              & w/           & 50.0   & 51.5   & 55.0   & 50.0   & 67.5  & 77.5 & 71.0                     & 72.5                     & 71.3 & 78.8 & 72.5   & 70.0   & 69.0   & 87.5   & 90.0  & 59.0    & 67.5    & 65.0    & 52.5    & 55.0   \\ \midrule
\multirow{11}{*}{Gemma3-12B}  & ROL          & w/o          & 39.5   & 39.5   & 41.5   & 40.0   & 41.0  & 71.0 & 70.0 & 72.5 & 72.0 & 73.5 & 60.0   & 57.5   & 59.5   & 59.5   & 55.0  & 40.0    & 44.0    & 42.5    & 43.5    & 43.0   \\
                              &              & w/           & 40.0   & 45.0   & 55.0   & 40.0   & 57.4  & 75.0 & 73.5                     & 75.0                     & 80.5 & 75.0 & 60.0   & 65.0   & 75.0   & 60.0   & 60.0  & 55.0    & 55.0    & 62.5    & 55.0    & 60.0   \\
                              & DIR          & w/o          & 60.5   & 63.0   & 48.5   & 60.0   & 56.5  & 74.0 & 77.5                     & 83.5                     & 79.5 & 100  & 70.5   & 67.0   & 64.5   & 67.0   & 82.2  & 51.0    & 51.0    & 47.0    & 50.0    & 67.5   \\
                              &              & w/           & 73.0   & 60.0   & 53.0   & 67.5   & 62.5  & 75.0 & 81.5                     & 85.0                     & 82.5 & 100  & 72.5   & 72.0   & 75.0   & 70.0   & 83.0  & 65.0    & 65.0    & 62.5    & 60.4    & 77.5   \\
                              & ADI          & w/o          & 41.0   & 40.0   & 44.0   & 42.0   & 44.0  & 74.5 & 73.5                     & 73.0                     & 73.0 & 73.0 & 59.5   & 57.4   & 62.0   & 71.5   & 88.5  & 43.5    & 44.0    & 42.5    & 44.5    & 44.0   \\
                              &              & w/           & 60.0   & 47.5   & 47.5   & 57.5   & 51.0  & 75.0 & 80.0                     & 85.0                     & 75.0 & 73.5 & 59.0   & 57.5   & 60.0   & 75.0   & 94.0  & 65.0    & 52.5    & 59.5    & 57.4    & 67.5   \\
                              & OFT          & w/o          & 40.5   & 43.0   & 38.5   & 39.5   & 40.0  & 72.0 & 72.0                     & 72.0                     & 73.0 & 72.0 & 57.5   & 56.0   & 55.5   & 59.5   & 54.5  & 42.0    & 42.5    & 43.5    & 43.5    & 44.0   \\
                              &              & w/           & 47.5   & 47.5   & 45.0   & 57.4   & 40.0  & 77.5 & 75.0                     & 82.5                     & 73.0 & 75.0 & 62.5   & 52.5   & 75.0   & 55.0   & 77.5  & 67.5    & 52.5    & 62.5    & 55.0    & 52.5   \\
                              & EXA          & w/o          & 40.5   & 40.5   & 42.0   & 41.5   & 44.5  & 72.0 & 73.0                     & 72.0                     & 71.5 & 71.5 & 60.0   & 53.5   & 56.5   & 59.5   & 81.5  & 42.0    & 43.5    & 45.0    & 43.5    & 44.5   \\
                              &              & w/           & 40.0   & 40.0   & 50.0   & 55.0   & 47.5  & 72.5 & 80.0                     & 82.5                     & 72.5 & 85.0 & 60.0   & 60.0   & 52.5   & 59.5   & 97.5  & 55.0    & 62.5    & 65.0    & 60.0    & 52.5   \\
                              & CRT          & w/o          & 40.5   & 40.5   & 41.5   & 43.5   & 45.0  & 73.5 & 73.5                     & 76.0                     & 72.0 & 77.5 & 59.5   & 62.5   & 61.0   & 68.5   & 64.0  & 46.0    & 44.0    & 45.0    & 44.0    & 49.5   \\
                              &              & w/           & 42.5   & 42.5   & 52.5   & 60.0   & 47.5  & 77.0 & 77.5                     & 80.0                     & 75.0 & 80.0 & 67.4   & 65.0   & 62.5   & 75.0   & 80.0  & 57.4    & 62.5    & 62.5    & 52.5    & 60.0   \\ \bottomrule
\end{tabular}
% } % <-- This line scales table to fit width
\\[1ex]
\caption{ASR (\%) on five LLMs across four datasets, grouped by prompt components and perturbation types. \textit{Note:} Prompt Components (COM): ROL = Role, DIR = Directive, ADI = Additional Information, OFT = Output Formatting, EXA = Examples. Perturbation Types: SCI = Special Character Insertion, SYR = Synonym Replacement, WOD = Word Deletion, SER = Sentence Rewriting, COD = Component Deletion. CRT = Control Group, AVG = Average Result.}

\label{tab:asr_components}
\end{table*}

Table~\ref{tab:asr_components} reports the ASRs of five LLMs across four domain-specific datasets, categorized by prompt components and perturbation types, revealing critical patterns in adversarial robustness. The DIR and ADI consistently yield higher ASRs, highlighting their centrality in conveying task-critical semantics. In contrast, the ROL shows greater resilience, likely due to its auxiliary function. Semantic perturbations, including SYR and SER, are generally more effective than syntactic ones like SCI, underscoring how meaning-level disruptions more severely compromise model behavior. Furthermore, model-specific vulnerabilities emerge, with \textit{LLaMA3-70B} and \textit{Gemma3-12B} being particularly susceptible to component-wise attacks, possibly due to architectural or training differences. Heatmaps in Figures~\ref{fig:heatmap-gpt} to \ref{fig:heatmap-gemma} (See in the Appendix \ref{sec:heatmap}) visualize these patterns, showing that datasets such as \textit{EMEA-PA} and \textit{Leetcode-PA} exhibit higher ASRs. These prompts depend heavily on semantically dense instructions and multi-step reasoning—especially within DIR and ADI components—making them more fragile under perturbation. By contrast, prompts from \textit{PubMedQA-PA} and \textit{CodeGeneration-PA}, which are more syntax-oriented or retrieval-based, display greater robustness, as superficial semantic alterations have less impact on task execution. These patterns underscore another observation:

\begin{tcolorbox}[
  colback=pink!15,
  colframe=black!40,
  boxrule=1pt,
  left=2pt, right=2pt, top=2pt, bottom=2pt,
  sharp corners,
  enhanced,
  drop shadow
]
\textbf{Finding \#2}: Prompt Components Exhibit Heterogeneous Robustness to Adversarial Perturbations.
\end{tcolorbox}

This leads to the following actionable insight for further improving robustness for prompt engineers and developers.

\begin{tcolorbox}[
  colback=blue!10!white,
  colframe=black!40,
  boxrule=1pt,
  left=2pt, right=2pt, top=2pt, bottom=2pt,
  sharp corners,
  enhanced,
  drop shadow
]
\textbf{Guideline \#2}: Prompt engineers should prioritize protecting high-impact components, while developers should focus defenses on semantically critical elements.
\end{tcolorbox}

\begin{figure}
    \centering
    \includegraphics[width=\linewidth]{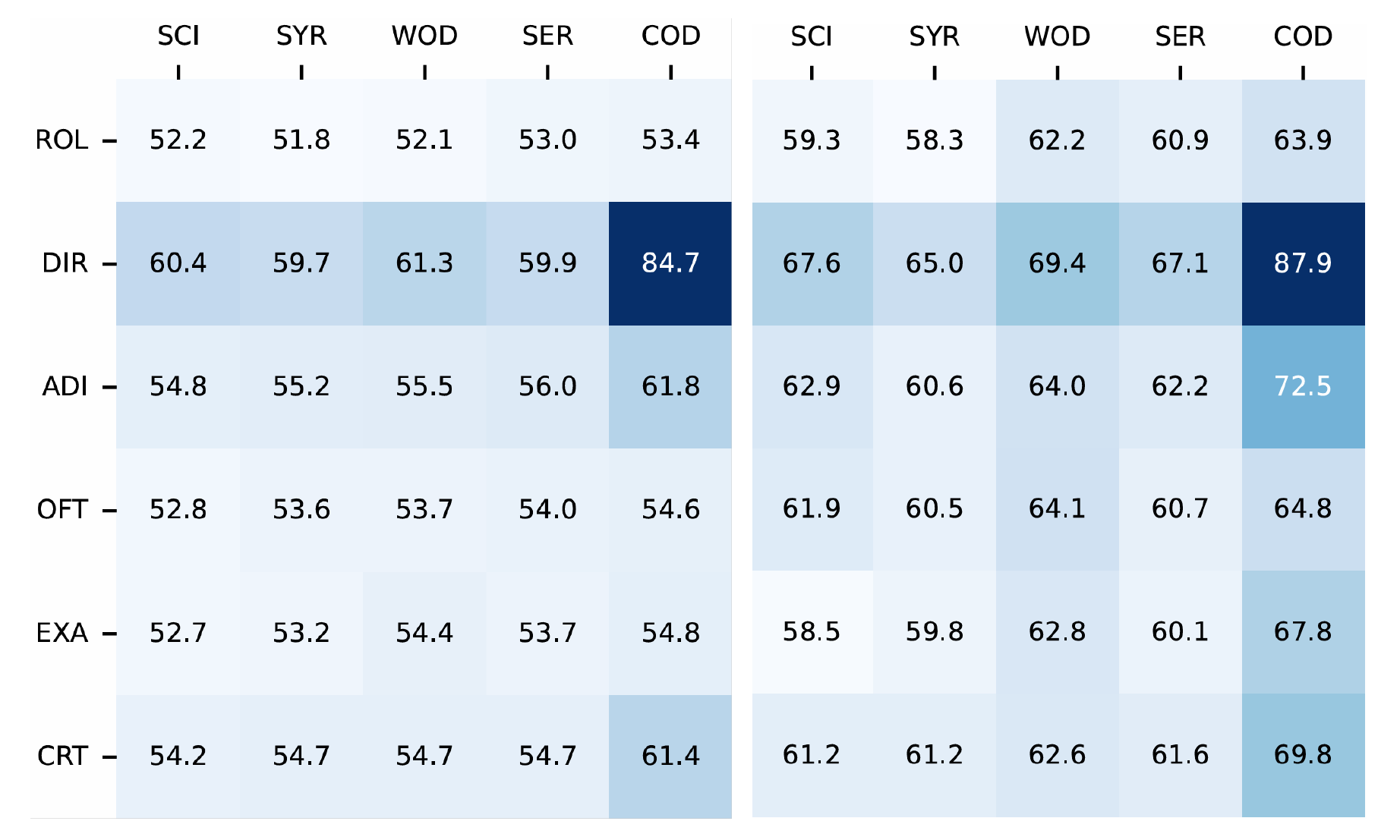}
    \caption{Heatmap of ASR across different prompt components and perturbation types. (left: w/o PPL, right: w/ PPL)}
    \label{fig:heatmap_component_perturbation}
\end{figure}

Figure~\ref{fig:heatmap_component_perturbation}, extracted from Table~\ref{tab:asr_components}, shows that attack success rates (ASRs) vary significantly depending on which prompt component is targeted and by which perturbation type. The DIR component consistently exhibits the highest ASRs—particularly under COD—reaching up to 87.9\%, underscoring its critical role in guiding model behavior. This heightened vulnerability stems from the semantic function of DIR, which conveys task-defining instructions; altering or removing it (e.g., via COD or SYR) disrupts task understanding. Similarly, ADI, which provides essential context or constraints, shows moderate susceptibility, especially to COD. In contrast, ROL and OFT serve more auxiliary purposes, such as setting tone or format, and thus display lower ASRs across perturbations. The EXA component yields moderate effects, suggesting limited impact in zero-shot settings. The Control group (CRT), where no semantically meaningful component is perturbed, yields the lowest ASRs, reinforcing that perturbation effectiveness is tightly coupled with a component’s functional importance. These results highlight the component-sensitive nature of LLMs, where disrupting core semantic components leads to disproportionately degraded performance, and also suggest that the type of perturbation—not just the target component—plays a key role in adversarial effectiveness:

\begin{tcolorbox}[
  colback=pink!15,
  colframe=black!40,
  boxrule=1pt,
  left=2pt, right=2pt, top=2pt, bottom=2pt,
  sharp corners,
  enhanced,
  drop shadow
]
\textbf{Finding \#3}: Semantic Perturbations Are More Effective Than Syntactic Ones Across Prompt Components.
\end{tcolorbox}

Building on this finding, we recommend the following best practice for both users and developers.

\begin{tcolorbox}[
  colback=blue!10!white,
  colframe=black!40,
  boxrule=1pt,
  left=2pt, right=2pt, top=2pt, bottom=2pt,
  sharp corners,
  enhanced,
  drop shadow
]
\textbf{Guideline \#3}: Users should avoid vague rephrasing, and developers should train models to withstand meaning-level changes for improved robustness.
\end{tcolorbox}

We conduct an ablation study to evaluate the impact of \textsc{PromptAnatomy} (PA) and \textsc{ComPerturb} (CP) across five LLMs, testing four configurations with and without each component. As shown in Table~\ref{tab:ablation_study}, using PA or CP individually boosts ASR by +9.2\% and +7.7\%, respectively, demonstrating their independent effectiveness. Combining both yields the highest gain, with a +15.4\% average ASR improvement, confirming their complementary strengths in enhancing attack performance.

\begin{table}[t]
\centering
\setlength{\tabcolsep}{1.5pt}
\renewcommand{\arraystretch}{1.1}
{\fontsize{8pt}{9.5pt}\selectfont
% \resizebox{\linewidth}{!}{ % <-- This line scales table to fit width
\begin{tabular}{cc|ccccc|c}
\toprule
\textbf{PA} & \textbf{CP} & \multicolumn{5}{c|}{\textbf{Attack Success Rate (\%)}} & \textbf{AVG $\uparrow$} \\
\cmidrule(lr){3-7}
& & \textbf{M1} & \textbf{M2} & \textbf{M3} & \textbf{M4} & \textbf{M5} & \\
\midrule
\ding{55} & \ding{55} & 55.9$\pm$0.12 & 54.1$\pm$0.08 & 54.6$\pm$0.15 & 58.9$\pm$0.15 & 56.4$\pm$0.15 & – \\
\ding{51} & \ding{55} & 64.8$\pm$0.11 & 62.8$\pm$0.09 & 64.5$\pm$0.13 & 67.8$\pm$0.13 & 66.0$\pm$0.13 & +9.2 \\
\ding{55} & \ding{51} & 64.0$\pm$0.09 & 61.8$\pm$0.06 & 60.8$\pm$0.13 & 66.7$\pm$0.11 & 65.0$\pm$0.11 & +7.7 \\
\ding{51} & \ding{51} & 72.3$\pm$0.08 & 70.3$\pm$0.06 & 69.4$\pm$0.11 & 72.8$\pm$0.09 & 72.0$\pm$0.10 & +15.4 \\
\bottomrule
\end{tabular}
}
\caption{Ablation study showing the effect of \textsc{PromptAnatomy} (PA) and \textsc{ComPerturb} (CP) on ASR.}
\label{tab:ablation_study}
\end{table}

\section{Discussion and Takeaways}

Our findings show that prompt components vary in adversarial robustness, challenging the notion that prompts are structurally uniform. (\romannumeral1) For security researchers, focusing on semantically critical components like Directive and Additional Information enables more effective red and blue teaming. (\romannumeral2) For prompt engineers and developers, adopting a component-aware design enhances robustness and control in high-stakes scenarios. (\romannumeral3) For general users, clearly specifying the Directive and relevant context improves model understanding and response quality.

\section{Conclusion}
We reveal that different prompt components exhibit heterogeneous adversarial vulnerabilities in large language models, challenging the prevalent assumption that prompts are uniformly robust. To facilitate fine-grained robustness analysis, we introduced \textsc{PromptAnatomy}, a novel framework that structurally dissects prompts into canonical components, and \textsc{ComPerturb}, a component-wise perturbation method. Extensive experiments show that vulnerabilities vary by component, with perturbation effectiveness tied to semantics and increasing with prompt complexity. These findings highlight the importance of structure-aware adversarial evaluation and the effectiveness of component-guided design in improving prompt robustness. We advocate for structure-aware prompt engineering and training-time defenses to build safer, more robust language models.

% Based on the component-wise nature of \textsc{PromptAnatomy}, we can explore a family of \textbf{targeted} adversarial attacks that manipulate individual prompt components. For example, attacks such as \texttt{DirectiveFlip}, \texttt{RoleInversion}, or \texttt{FormatJammer} could be defined to selectively perturb the \texttt{Directive}, \texttt{Role}, or \texttt{OutputFormat} components, respectively. This direction could lead to a more interpretable and controllable adversarial evaluation framework, enabling fine-grained robustness analysis across prompt structures.

\clearpage
\bibliography{aaai2026}

\onecolumn
\section{Appendix}

\subsection{Examples of LLM Fails to Dissect Prompt}\label{sec:llm_fails_to_decompose}

\paragraph{Misdissection of Semantic Components by GPT-4o}

\begin{tcolorbox}[title=, breakable]
\textbf{Prompt to GPT-4o}\\
You are now an expert in prompt engineering for large language models. Please classify the following prompt into the five components: Role, Directive, Additional Information, Output Formatting and Examples. Please classify the following prompt into its respective components using implicit tags such as \texttt{<Role></Role>}, \texttt{<Directive></Directive>}, \texttt{<Additional Information></Additional Information>}, \texttt{<Output Formatting></Output Formatting>} and \texttt{<Examples></Examples>}. Do not provide any explanations or additional content. Each sentence should remain unmodified. The classified sentences should remain the same as the original sentences. Below are the definitions of each component: Role: Also known as a persona, this component helps improve the writing style and tone of the text. Examples: As a helpful assistant, answer the following question concisely. Directive: The main instruction or question in the prompt. i.e. What are the benefits of renewable energy? Or Tell me five books. Additional Information: Provides additional background information or situational details to help the model better understand the task. Example: You are writing an email to a colleague about a meeting next week. Output Formatting: Defines how the response should be structured or presented. Example: Answer the question in 'yes' or 'no'. Examples: Provides sample inputs and outputs to guide the model's response. Example: Input: 'Night' Output: 'Noche' (Spanish). The prompt to classify is:\\
A seasoned Python developer like you has extensive experience in writing and refining Python code, and your expertise includes key domains such as software engineering, data processing, and algorithm optimization. Your task is to either create new Python scripts that fulfill specific functional requirements or modify existing Python code to enhance its performance, readability, or functionality. These tasks involve implementing basic functions, optimizing existing code, and ensuring that the code adheres to best practices in software development. The output should be a well-structured Python script. It should include clear comments explaining the purpose of each section of the code. The script should be formatted according to PEP 8 guidelines, ensuring readability and consistency. Below is an instruction that describes a task, Write a response that appropriately completes the request. Inspect the following Python file from murano/tests/unit/api/middleware/test\_version\_negotiation.py. Update the given code considering this change: "Change assertTrue(isinstance()) by optimal assert" ```python \# Copyright 2016 AT\&T Corp \# All Rights Reserved.\# \# Licensed under the Apache License, Version 2.0 (the "License"); you may \# not use this file except in compliance with the License.You may obtain \# a copy of the License at \# \# http://www.apache.org/licenses/LICENSE-2.0 \# \# Unless required by applicable law or agreed to in writing, software \# distributed under the License is distributed on an "AS IS" BASIS, WITHOUT \# WARRANTIES OR CONDITIONS OF ANY KIND, either express or implied. See the \# License for the specific language governing permissions and limitations \# under the License.import webob from murano.api import versions from murano.api.middleware import version\_negotiation from murano.tests.unit import base class MiddlewareVersionNegotiationTest(base.MuranoTestCase): def test\_middleware\_version\_negotiation\_default(self): middleware\_vn = version\_negotiation.VersionNegotiationFilter(None) request = webob.Request.blank('/environments') result = middleware\_vn.process\_request(request) self.assertTrue(isinstance(result, versions.Controller)) ```. Example : Input: Generate a program in Python that takes in a string and returns the reverse of the string."Hello World" Output: def reverse\_string(str): """ Reverses a given string """ rev\_str = "" for c in str: rev\_str = c + rev\_str return rev\_str str = "Hello World" rev\_str = reverse\_string(str) print(rev\_str).
\\

\textbf{Ground Truth}\\
\texttt{<Role>} A seasoned Python developer like you has extensive experience in writing and refining Python code, and your expertise includes key domains such as software engineering, data processing, and algorithm optimization.\texttt{</Role>}\\
\texttt{<Directive>} Your task is to either create new Python scripts that fulfill specific functional requirements or modify existing Python code to enhance its performance, readability, or functionality. Below is an instruction that describes a task, Write a response that appropriately completes the request. Inspect the following Python file from murano/tests/unit/api/middleware/test\_version\_negotiation.py. Update the given code considering this change: "Change assertTrue(isinstance()) by optimal assert" ```python \# Copyright 2016 AT\&T Corp \# All Rights Reserved.\# \# Licensed under the Apache License, Version 2.0 (the "License"); you may \# not use this file except in compliance with the License. You may obtain \# a copy of the License at \# \# http://www.apache.org/licenses/LICENSE-2.0 \# \# Unless required by applicable law or agreed to in writing, software \# distributed under the License is distributed on an "AS IS" BASIS, WITHOUT \# WARRANTIES OR CONDITIONS OF ANY KIND, either express or implied. See the \# License for the specific language governing permissions and limitations \# under the License. import webob from murano.api import versions from murano.api.middleware import version\_negotiation from murano.tests.unit import base class MiddlewareVersionNegotiationTest(base.MuranoTestCase): def test\_middleware\_version\_negotiation\_default(self): middleware\_vn = version\_negotiation.VersionNegotiationFilter(None) request = webob.Request.blank('/environments') result = middleware\_vn.process\_request(request) self.assertTrue(isinstance(result, versions.Controller)) ```. \texttt{</Directive>}\\
\texttt{<Additional Information>} These tasks involve implementing basic functions, optimizing existing code, and ensuring that the code adheres to best practices in software development.\texttt{</Additional Information>}\\
\texttt{<Output Formatting>} The output should be a well-structured Python script. It should include clear comments explaining the purpose of each section of the code. The script should be formatted according to PEP 8 guidelines, ensuring readability and consistency.\texttt{</Output Formatting>}\\
\texttt{<Examples>}Example : Input: Generate a program in Python that takes in a string and returns the reverse of the string."Hello World" Output: def reverse\_string(str): """ Reverses a given string """ rev\_str = "" for c in str: rev\_str = c + rev\_str return rev\_str str = "Hello World" rev\_str = reverse\_string(str) print(rev\_str).\texttt{</Examples>}\\

\textbf{Response From GPT-4o}\\
\texttt{<Role>} A seasoned Python developer like you has extensive experience in writing and refining Python code, and your expertise includes key domains such as software engineering, data processing, and algorithm optimization. \texttt{</Role>}

\texttt{<Directive>}Your task is to either create new Python scripts that fulfill specific functional requirements or modify existing Python code to enhance its performance, readability, or functionality. \texttt{</Directive>}

\texttt{<Additional Information>}These tasks involve implementing basic functions, optimizing existing code, and ensuring that the code adheres to best practices in software development. \texttt{</Additional Information>}

\texttt{<Output Formatting>}The output should be a well-structured Python script. It should include clear comments explaining the purpose of each section of the code. The script should be formatted according to PEP 8 guidelines, ensuring readability and consistency. \texttt{</Output Formatting>}

\textcolor{red}{\texttt{<Additional Information>} Below is an instruction that describes a task, Write a response that appropriately completes the request. \texttt{</Additional Information>} }

\texttt{<Examples>} \textcolor{red}{Inspect the following Python file from murano/tests/unit/api/middleware/test\_version\_negotiation.py. Update the given code considering this change: "Change assertTrue(isinstance()) by optimal assert" ```python \# Copyright 2016 AT\&T Corp \# All Rights Reserved.\# \# Licensed under the Apache License, Version 2.0 (the }\textcolor{red}{"Li}\textcolor{red}{cense"); you may \# not use this file except in compliance with the License.You may obtain \# a copy of the License at \# \# http://www.apache.org/licenses/LICENSE-2.0 \# \# Unless required by applicable law or agreed to in writing, software \# distributed under the License is distributed on an "AS IS" BASIS, WITHOUT \# WARRANTIES OR CONDITIONS OF ANY KIND, either express or implied.See the \# License for the specific language governing permissions and limitations \# under the License.import webob from murano.api import versions from murano.api.middleware import version\_negotiation from murano.tests.unit import base class MiddlewareVersionNegotiationTest(base.MuranoTestCase): def test\_middleware\_version\_negotiation\_default(self): middleware\_vn = version\_negotiation.VersionNegotiationFilter(None) request = webob.Request.blank('/environments') result = middleware\_vn.process\_request(request) self.assertTrue(isinstance(result, versions.Controller)) ```.} Example : Input: Generate a program in Python that takes in a string and returns the reverse of the string."Hello World" Output: def reverse\_string(str): """ Reverses a given string """ rev\_str = "" for c in str: rev\_str = c + rev\_str return rev\_str str = "Hello World" rev\_str = reverse\_string(str) print(rev\_str) \texttt{</Examples>}

\end{tcolorbox}

\paragraph{Omission of Middle Sentences in Prompt Dissection by GPT-4o}
\begin{tcolorbox}[title=, breakable]
\textbf{Prompt to GPT-4o}\\
You are now an expert in prompt engineering for large language models. Please classify the following prompt into the five components: Role, Directive, Additional Information, Output Formatting and Examples. Please classify the following prompt into its respective components using implicit tags such as \texttt{<Role></Role>}, \texttt{<Directive></Directive>}, \texttt{<Additional Information></Additional Information>}, \texttt{<Output Formatting></Output Formatting>} and \texttt{<Examples></Examples>}. Do not provide any explanations or additional content. Each sentence should remain unmodified. The classified sentences should remain the same as the original sentences. Below are the definitions of each component: Role: Also known as a persona, this component helps improve the writing style and tone of the text. Examples: As a helpful assistant, answer the following question concisely. Directive: The main instruction or question in the prompt. i.e. What are the benefits of renewable energy? Or Tell me five books. Additional Information: Provides additional background information or situational details to help the model better understand the task. Example: You are writing an email to a colleague about a meeting next week. Output Formatting: Defines how the response should be structured or presented. Example: Answer the question in 'yes' or 'no'. Examples: Provides sample inputs and outputs to guide the model's response. Example: Input: 'Night' Output: 'Noche' (Spanish). The prompt to classify is:\\
As a cybersecurity expert dedicated to password strength analysis and improvement, you are responsible for assessing and adjusting passwords to meet stringent security criteria. You are given a password and you need to generate the number of steps required to convert the given password to a strong password. A password is considered strong if (a) it has at least 6 characters and at most 20 characters; (b) it contains at least one lowercase letter and one uppercase letter, and at least one digit; (c) it does not contain three repeating characters in a row. In one step you can: (1) Insert one character to password, (2) delete one character from password, or (3) replace one character of password with another character. Calculate the minimum number of operations needed to transform the provided password into a strong one. Make sure your approach is efficient and explores all potential scenarios for optimal results. Answer only returns the number of steps required. Now complete the question: password = KMxgCvxC4A1SxLS2r9OEQGx3MqRaqXX. Here are some examples: Positive Example 1 - Input: password = a Output: 5.Positive Example 2 - Input: password = aA1 Output: 3.Negative Example 1 - Input: password = 1337C0d3 Output: 1.\\

\textbf{Ground Truth}\\
\texttt{<Role>}As a cybersecurity expert dedicated to password strength analysis and improvement, you are responsible for assessing and adjusting passwords to meet stringent security criteria.\texttt{</Role>}\\
\texttt{<Directive>}You are given a password and you need to generate the number of steps required to convert the given password to a strong password. Calculate the minimum number of operations needed to transform the provided password into a strong one. Now complete the question: password = KMxgCvxC4A1SxLS2r9OEQGx3MqRaqXX.\texttt{</Directive>}\\
\texttt{<Additional Information>}A password is considered strong if (a) it has at least 6 characters and at most 20 characters; (b) it contains at least one lowercase letter and one uppercase letter, and at least one digit; (c) it does not contain three repeating characters in a row. In one step you can: (1) Insert one character to password, (2) delete one character from password, or (3) replace one character of password with another character. Make sure your approach is efficient and explores all potential scenarios for optimal results.\texttt{</Additional Information>}\\
\texttt{<Output Formatting>}Answer only returns the number of steps required.\texttt{</Output Formatting>}\\
\texttt{<Examples>}Here are some examples: Positive Example 1 - Input: password = a Output: 5.Positive Example 2 - Input: password = aA1 Output: 3.Negative Example 1 - Input: password = 1337C0d3 Output: 1.\texttt{</Examples>}\\

\textbf{Response From GPT-4o}\\
\texttt{<Role>} As a cybersecurity expert dedicated to password strength analysis and improvement, you are responsible for assessing and adjusting passwords to meet stringent security criteria. \texttt{</Role>}\\
\texttt{<Directive>} You are given a password and you need to generate the number of steps required to convert the given password to a strong password. \texttt{</Directive>}\\
\texttt{<Additional Information>} A password is considered strong if (a) it has at least 6 characters and at most 20 characters; (b) it contains at least one lowercase letter and one uppercase letter, and at least one digit; (c) it does not contain three repeating characters in a row. In one step you can: (1) Insert one character to password, (2) delete one character from password, or (3) replace one character of password with another character. Make sure your approach is efficient and explores all potential scenarios for optimal results. \texttt{</Additional Information>}\\
\texttt{<Output Formatting>} Answer only returns the number of steps required. \texttt{</Output Formatting>}\\
\texttt{<Examples>} Positive Example 1 - Input: password = a Output: 5. Positive Example 2 - Input: password = aA1 Output: 3. Negative Example 1 - Input: password = 1337C0d3 Output: 1. \texttt{</Examples>}\\
\textcolor{red}{Missing Sentence:}\\
\textcolor{red}{Calculate the minimum number of operations needed to transform the provided password into a strong one.}\\
\textcolor{red}{Now complete the question: password = KMxgCvxC4A1SxLS2r9OEQGx3MqRaqXX.}

\end{tcolorbox}

\newpage
\subsection{Examples of Origin and Enhanced Datasets}\label{sec:datasets}

\begin{longtable}{p{2.5cm}p{13cm}}
\caption{Dataset Examples from PubMedQA, EMEA, Leetcode, and CodeGeneration} \label{tab:dataset_samples} \\
\toprule
Dataset & Example \\ \hline
\endfirsthead

\toprule
Dataset & Example \\ \hline
\endhead

\hline
\endfoot

\hline
\endlastfoot

PubMedQA &
You are a response inspector, ensuring that the answer is a valid reply to the question. Check if the answer to a question is correct and classify the result as 'yes' or 'no'. Output 'yes' if the answer answers the question and output 'no' if the answer does not answer the question. Focus on determining if the answer is both relevant and complete in addressing the question. Now complete the question: Question: Is chk1 required for the metaphase-anaphase transition via regulating the expression and localization of Cdc20 and Mad2? Answer: These results strongly suggest that Chk1 is required for the metaphase-anaphase transition via regulating the subcellular localization and the expression of Cdc20 and Mad2. Here are some examples: Positive Example 1 - Input: Question: Are group 2 innate lymphoid cells (ILC2s) increased in chronic rhinosinusitis with nasal polyps or eosinophilia? Answer: As ILC2s are elevated in patients with CRSwNP, they may drive nasal polyp formation in CRS. ILC2s are also linked with high tissue and blood eosinophilia and have a potential role in the activation and survival of eosinophils during the Th2 immune response. The association of innate lymphoid cells in CRS provides insights into its pathogenesis. Output: yes Positive Example 2 - Input: Question: Does vagus nerve contribute to the development of steatohepatitis and obesity in phosphatidylethanolamine N-methyltransferase deficient mice? Answer: Neuronal signals via the hepatic vagus nerve contribute to the development of steatohepatitis and protection against obesity in HFD fed Pemt(-/-) mice. Output: yes Negative Example 1 - Input: Question: Is methylation of the FGFR2 gene associated with high birth weight centile in humans? Answer: We identified a novel biologically plausible candidate (FGFR2) for with BWC that merits further study. Output: no Negative Example 2 - Input: Question: Do tumor-infiltrating immune cell profiles and their change after neoadjuvant chemotherapy predict response and prognosis of breast cancer? Answer: Breast cancer immune cell subpopulation profiles, determined by immunohistochemistry-based computerized analysis, identify groups of patients characterized by high response (in the pre-treatment setting) and poor prognosis (in the post-treatment setting). Further understanding of the mechanisms underlying the distribution of immune cells and their changes after chemotherapy may contribute to the development of new immune-targeted therapies for breast cancer. Output: no. \\ \hline

EMEA &
As an experienced Latin translation specialist, you have a deep knowledge of historical and religious contexts, which enables you to translate Latin texts into English with precision. Your skill lies in grasping the subtleties of classical languages and their cultural importance. Translate the provided Latin text into English, ensuring that the historical and religious nuances are preserved and accurately conveyed. The content you've provided is extensive, with a wide range of texts and references from various fields like religion, history, philosophy, and literature. The translation should be presented in a clear and readable English format, with any necessary annotations or explanations provided to clarify the context or meaning of specific terms or phrases. Now complete the question: ceterum cum et magni pretii et varii generis a legatis eius tam virorum quam feminarum apta usui munera circa domos ferrentur, nulla cuiquam dono ianua patuit, Tarentinaeque petulantiae animosus magis quam efficax defensor haud scio maiore cum gloria huius urbis moribus <an moenibus> repulsus sit. Here are some examples: Latin Text: Omnia mutantur, nihil interit. Translation: Everything changes, nothing perishes. Latin Text: Qui cum statuisset, nisi me per vos recuperasset, eamdem subire fortunam atque idem sibi domicilium et vitae et mortis deposcere, tamen numquam nec magnitudinem negotii nec solitudinem suam nec vim inimicorum ac tela pertimuit. Translation: He had made up his mind that, should he fail, through you, to win me back to himself, he would ask permission to meet the same fate and to share the same dwelling with me in life and in death; and yet, in spite of this, no toil however formidable, no loneliness, no threat nor weapons of foes, could daunt him. \\ \hline

Leetcode &
As a cybersecurity expert dedicated to password strength analysis and improvement, you are responsible for assessing and adjusting passwords to meet stringent security criteria. You are given a password and you need to generate the number of steps required to convert the given password to a strong password. A password is considered strong if (a) it has at least 6 characters and at most 20 characters; (b) it contains at least one lowercase letter and one uppercase letter, and at least one digit; (c) it does not contain three repeating characters in a row. In one step you can: (1) Insert one character to password, (2) delete one character from password, or (3) replace one character of password with another character. Calculate the minimum number of operations needed to transform the provided password into a strong one. Make sure your approach is efficient and explores all potential scenarios for optimal results. Answer only returns the number of steps required. Now complete the question: password = KMxgCvxC4A1SxLS2r9OEQGx3MqRaqXX. Here are some examples: Positive Example 1 - Input: password = a Output: 5. Positive Example 2 - Input: password = aA1 Output: 3. Negative Example 1 - Input: password = 1337C0d3 Output: 1. \\ \hline

CodeGeneration &
A seasoned Python developer like you has extensive experience in writing and refining Python code, and your expertise includes key domains such as software engineering, data processing, and algorithm optimization. Your task is to either create new Python scripts that fulfill specific functional requirements or modify existing Python code to enhance its performance, readability, or functionality. These tasks involve implementing basic functions, optimizing existing code, and ensuring that the code adheres to best practices in software development. The output should be a well-structured Python script. It should include clear comments explaining the purpose of each section of the code. The script should be formatted according to PEP 8 guidelines, ensuring readability and consistency. Below is an instruction that describes a task, Write a response that appropriately completes the request. Inspect the following Python file from murano/tests/unit/api/middleware/test\_version\_negotiation.py. Update the given code considering this change: "Change assertTrue(isinstance()) by optimal assert" ```python \# Copyright 2016 AT\&T Corp \# All Rights Reserved.\# \# Licensed under the Apache License, Version 2.0 (the "License"); you may \# not use this file except in compliance with the License.You may obtain \# a copy of the License at \# \# http://www.apache.org/licenses/LICENSE-2.0 \# \# Unless required by applicable law or agreed to in writing, software \# distributed under the License is distributed on an "AS IS" BASIS, WITHOUT \# WARRANTIES OR CONDITIONS OF ANY KIND, either express or implied. See the \# License for the specific language governing permissions and limitations \# under the License.import webob from murano.api import versions from murano.api.middleware import version\_negotiation from murano.tests.unit import base class MiddlewareVersionNegotiationTest(base.MuranoTestCase): def test\_middleware\_version\_negotiation\_default(self): middleware\_vn = version\_negotiation.VersionNegotiationFilter(None) request = webob.Request.blank('/environments') result = middleware\_vn.process\_request(request) self.assertTrue(isinstance(result, versions.Controller)) ```. Example : Input: Generate a program in Python that takes in a string and returns the reverse of the string."Hello World" Output: def reverse\_string(str): """ Reverses a given string """ rev\_str = "" for c in str: rev\_str = c + rev\_str return rev\_str str = "Hello World" rev\_str = reverse\_string(str) print(rev\_str).  \\ \hline
\end{longtable}

\begin{longtable}{p{2.5cm}p{13cm}}
\caption{Dataset Examples from PubMedQA-PA, EMEA-PA, Leetcode-PA, and CodeGeneration-PA} \label{tab:dataset_samples-PA} \\
\toprule
Dataset & Example \\ \hline
\endfirsthead

\toprule
Dataset & Example \\ \hline
\endhead

\hline
\endfoot

\hline
\endlastfoot

PubMedQA-PA &
\texttt{<Role>}You are a response inspector, ensuring that the answer is a valid reply to the question.\texttt{</Role>} 
\texttt{<Directive>}Check if the answer to a question is correct and classify the result as 'yes' or 'no'. Now complete the question: Question: Is chk1 required for the metaphase-anaphase transition via regulating the expression and localization of Cdc20 and Mad2? Answer: These results strongly suggest that Chk1 is required for the metaphase-anaphase transition via regulating the subcellular localization and the expression of Cdc20 and Mad2. \texttt{</Directive>}
\texttt{<Additional Information>}Focus on determining if the answer is both relevant and complete in addressing the question.\texttt{</Additional Information>}
\texttt{<Output Formatting>}Output 'yes' if the answer answers the question and output 'no' if the answer does not answer the question. \texttt{</Output Formatting> }
\texttt{<Examples>}Here are some examples: Positive Example 1 - Input: Question: Are group 2 innate lymphoid cells (ILC2s) increased in chronic rhinosinusitis with nasal polyps or eosinophilia? Answer: As ILC2s are elevated in patients with CRSwNP, they may drive nasal polyp formation in CRS. ILC2s are also linked with high tissue and blood eosinophilia and have a potential role in the activation and survival of eosinophils during the Th2 immune response. The association of innate lymphoid cells in CRS provides insights into its pathogenesis. Output: yes Positive Example 2 - Input: Question: Does vagus nerve contribute to the development of steatohepatitis and obesity in phosphatidylethanolamine N-methyltransferase deficient mice? Answer: Neuronal signals via the hepatic vagus nerve contribute to the development of steatohepatitis and protection against obesity in HFD fed Pemt(-/-) mice. Output: yes Negative Example 1 - Input: Question: Is methylation of the FGFR2 gene associated with high birth weight centile in humans? Answer: We identified a novel biologically plausible candidate (FGFR2) for with BWC that merits further study. Output: no Negative Example 2 - Input: Question: Do tumor-infiltrating immune cell profiles and their change after neoadjuvant chemotherapy predict response and prognosis of breast cancer? Answer: Breast cancer immune cell subpopulation profiles, determined by immunohistochemistry-based computerized analysis, identify groups of patients characterized by high response (in the pre-treatment setting) and poor prognosis (in the post-treatment setting). Further understanding of the mechanisms underlying the distribution of immune cells and their changes after chemotherapy may contribute to the development of new immune-targeted therapies for breast cancer. Output: no.\texttt{</Examples>} \\ \hline

EMEA-PA &
\texttt{<Role>}As an experienced Latin translation specialist, you have a deep knowledge of historical and religious contexts, which enables you to translate Latin texts into English with precision.\texttt{</Role>} 
\texttt{<Directive>}Translate the provided Latin text into English, ensuring that the historical and religious nuances are preserved and accurately conveyed. Now complete the question: ceterum cum et magni pretii et varii generis a legatis eius tam virorum quam feminarum apta usui munera circa domos ferrentur, nulla cuiquam dono ianua patuit, Tarentinaeque petulantiae animosus magis quam efficax defensor haud scio maiore cum gloria huius urbis moribus <an moenibus> repulsus sit.\texttt{</Directive>}
\texttt{<Additional Information>}Your skill lies in grasping the subtleties of classical languages and their cultural importance.  The content you've provided is extensive, with a wide range of texts and references from various fields like religion, history, philosophy, and literature.\texttt{</Additional Information>} 
\texttt{<Output Formatting>}The translation should be presented in a clear and readable English format, with any necessary annotations or explanations provided to clarify the context or meaning of specific terms or phrases.\texttt{</Output Formatting>} 
\texttt{<Examples>}Here are some examples: Latin Text: Omnia mutantur, nihil interit. Translation: Everything changes, nothing perishes. Latin Text: Qui cum statuisset, nisi me per vos recuperasset, eamdem subire fortunam atque idem sibi domicilium et vitae et mortis deposcere, tamen numquam nec magnitudinem negotii nec solitudinem suam nec vim inimicorum ac tela pertimuit. Translation: He had made up his mind that, should he fail, through you, to win me back to himself, he would ask permission to meet the same fate and to share the same dwelling with me in life and in death; and yet, in spite of this, no toil however formidable, no loneliness, no threat nor weapons of foes, could daunt him.\texttt{</Examples>} \\ \hline

Leetcode-PA &
\texttt{<Role>}As a cybersecurity expert dedicated to password strength analysis and improvement, you are responsible for assessing and adjusting passwords to meet stringent security criteria. \texttt{</Role>}
\texttt{<Directive>}You are given a password and you need to generate the number of steps required to convert the given password to a strong password.Now complete the question: password = KMxgCvxC4A1SxLS2r9OEQGx3MqRaqXX.\texttt{</Directive>} 
\texttt{<Additional Information>}A password is considered strong if (a) it has at least 6 characters and at most 20 characters; (b) it contains at least one lowercase letter and one uppercase letter, and at least one digit; (c) it does not contain three repeating characters in a row. In one step you can: (1) Insert one character to password, (2) delete one character from password, or (3) replace one character of password with another character.Calculate the minimum number of operations needed to transform the provided password into a strong one. Make sure your approach is efficient and explores all potential scenarios for optimal results.\texttt{</Additional Information>}
\texttt{<Output Formatting>}Answer only returns the number of steps required.\texttt{</Output Formatting>}  
\texttt{<Examples>}Here are some examples: Positive Example 1 - Input: password = a Output: 5. Positive Example 2 - Input: password = aA1 Output: 3. Negative Example 1 - Input: password = 1337C0d3 Output: 1.\texttt{</Examples>} \\ \hline

CodeGeneration-PA &
\texttt{<Role>}A seasoned Python developer like you has extensive experience in writing and refining Python code, and your expertise includes key domains such as software engineering, data processing, and algorithm optimization.\texttt{</Role>} 
\texttt{<Directive>}Your task is to either create new Python scripts that fulfill specific functional requirements or modify existing Python code to enhance its performance, readability, or functionality. Below is an instruction that describes a task, Write a response that appropriately completes the request. Inspect the following Python file from murano/tests/unit/api/middleware/test\_version\_negotiation.py. Update the given code considering this change: "Change assertTrue(isinstance()) by optimal assert" ```python \# Copyright 2016 AT\&T Corp \# All Rights Reserved.\# \# Licensed under the Apache License, Version 2.0 (the "License"); you may \# not use this file except in compliance with the License.You may obtain \# a copy of the License at \# \# http://www.apache.org/licenses/LICENSE-2.0 \# \# Unless required by applicable law or agreed to in writing, software \# distributed under the License is distributed on an "AS IS" BASIS, WITHOUT \# WARRANTIES OR CONDITIONS OF ANY KIND, either express or implied. See the \# License for the specific language governing permissions and limitations \# under the License.import webob from murano.api import versions from murano.api.middleware import version\_negotiation from murano.tests.unit import base class MiddlewareVersionNegotiationTest(base.MuranoTestCase): def test\_middleware\_version\_negotiation\_default(self): middleware\_vn = version\_negotiation.VersionNegotiationFilter(None) request = webob.Request.blank('/environments') result = middleware\_vn.process\_request(request) self.assertTrue(isinstance(result, versions.Controller)) ```.\texttt{</Directive>} 
\texttt{<Additional Information>}These tasks involve implementing basic functions, optimizing existing code, and ensuring that the code adheres to best practices in software development.\texttt{</Additional Information>} 
\texttt{<Output Formatting>}The output should be a well-structured Python script. It should include clear comments explaining the purpose of each section of the code. The script should be formatted according to PEP 8 guidelines, ensuring readability and consistency. \texttt{</Output Formatting>}
\texttt{<Examples>}Example : Input: Generate a program in Python that takes in a string and returns the reverse of the string."Hello World" Output: def reverse\_string(str): """ Reverses a given string """ rev\_str = "" for c in str: rev\_str = c + rev\_str return rev\_str str = "Hello World" rev\_str = reverse\_string(str) print(rev\_str).\texttt{</Examples>}  \\ \bottomrule

\end{longtable}

\newpage
\subsection{Details of Prompt Perturbation Strategies}

To evaluate the robustness of LLMs under structurally and semantically aligned adversarial prompts, we design perturbations at four linguistic granularity levels. Each strategy is applied to individual semantic components in prompts. Table \ref{tab:example_per} is an example of five perturbation types applied to the \texttt{Role} component in prompts.

\paragraph{1) Character-Level Perturbation.}
Simulates typos or symbol noise. For each semantic component, we randomly insert special characters (e.g., \#, \$, \%, \&, @) at 10\% of token positions, evenly distributed across the span.\\
\textit{Original:} a sentiment classification agent.\\
\textit{Perturbed:} a senti!ment c\#lass@ification agent.

\paragraph{2) Word-Level Perturbation (Deletion).}
To mimic incomplete or informal text, 10\% of words in each component are randomly deleted.\\
\textit{Original:} a sentiment classification agent.\\
\textit{Perturbed:} a classification agent.

\paragraph{3) Word-Level Perturbation (Synonym Substitution).}
We substitute 10\% of words with synonyms from WordNet to simulate paraphrasing.\\
\textit{Original:} a sentiment classification agent.\\
\textit{Perturbed:} a sentiment analysis agent.

\paragraph{4) Sentence-Level Perturbation.}
One sentence per component is randomly replaced with a semantically equivalent sentence from a paraphrase corpus.\\
\textit{Original:} a sentiment analysis agent.\\
\textit{Perturbed:} a system designed for sentiment categorization.

\paragraph{5) Component-Level Perturbation.}
An entire semantic component (e.g., \texttt{<Role>}, \texttt{<Directive>}, \texttt{<Examples>}) is removed to assess its marginal contribution.\\
\textit{Example (original):}  
\texttt{<Role>...</Role> <Directive>...</Directive> <Examples>...</Examples>}  

\textit{Example (perturbed):}  
\texttt{<Directive>...</Directive> <Examples>...</Examples>} (Role component removed)

\begin{table}[h]
\centering
\caption{Examples of five perturbation types applied to the \texttt{Role} component in prompts. Each example shows how a clean prompt is modified using one specific perturbation strategy: Special Character Insertion (SCI), Synonym Replacement (SYR), Word Deletion (WOD), Sentence Rewriting (SER), and Component Deletion (COD).}
\label{tab:example_per}
\begin{tabular}{@{}ll@{}}
\toprule
Scenario                          & Prompt                                                                                                                                                                    \\ \midrule
Clean                             & \begin{tabular}[c]{@{}l@{}}You are Assistant, a sentiment classification agent. You will identify the sentiment \\ given an utterance or text.\end{tabular}               \\
Special Character Insertion (SCI) & \begin{tabular}[c]{@{}l@{}}You are Ass\textcolor{red}{!}istant, a sentiment c\textcolor{red}{@}lassification agent. You will identify the sentiment \\ given an utterance or text.\end{tabular}               \\
Synonym Replacement (SYR)         & \begin{tabular}[c]{@{}l@{}}You are Assistant, a  sentiment \textcolor{red}{analysis} agent.  You will identify the sentiment \\ given an utterance or text.\end{tabular}                \\
Word Deletion (WOD)               & \begin{tabular}[c]{@{}l@{}}You are Assistant, a sentiment classification.  You will identify the sentiment \\ given an utterance or text.\end{tabular}                     \\
Sentence Rewriting (SER)          & \begin{tabular}[c]{@{}l@{}}\textcolor{red}{You are Assistant, a system designed for sentiment categorization.}  You will identify the \\ sentiment given an utterance or text.\end{tabular} \\
Component Deletion (COD)          & \begin{tabular}[c]{@{}l@{}}You will identify the sentiment given an utterance or text.\end{tabular}               \\ \bottomrule
\end{tabular}
\end{table}

\newpage
\subsection{Prompt Complexity Metric}\label{sec:pcm}

This appendix details the Prompt Complexity Metric (PCM), which systematically quantifies the complexity of prompts used with Large Language Models (LLMs). The metric integrates lexical, syntactic, semantic, structural, and task-oriented dimensions into a unified measure.

\subsubsection{Formulation}

The PCM for a prompt $p$ is defined as:
\begin{equation}
\text{PCM}(p) = \alpha C_{\text{lexical}} + \beta C_{\text{syntactic}} + \gamma C_{\text{semantic}} + \delta C_{\text{structural}} + \epsilon C_{\text{task}}
\end{equation}
where coefficients $\alpha, \beta, \gamma, \delta, \epsilon$ are tunable parameters.

\subsubsection{Components}

\paragraph{Lexical Complexity ($C_{\text{lexical}}$)} Measures vocabulary complexity based on token rarity:
\begin{equation}
C_{\text{lexical}}(p) = \frac{1}{|p|}\sum_{w \in p}\text{IDF}(w)
\end{equation}
where $\text{IDF}(w)$ is the inverse document frequency of token $w$.

\paragraph{Syntactic Complexity ($C_{\text{syntactic}}$)} Represents sentence structure intricacy:
\begin{equation}
C_{\text{syntactic}}(p) = \frac{1}{|S|}\sum_{s \in S}\text{depth}(\text{ParseTree}(s))
\end{equation}
where $S$ is the set of sentences in $p$.

\paragraph{Semantic Complexity ($C_{\text{semantic}}$)} Measures semantic dispersion using embeddings:
\begin{equation}
C_{\text{semantic}}(p) = \frac{2}{|S|(|S|-1)}\sum_{s_i, s_j \in S; i<j}1 - \text{cos}(E(s_i), E(s_j))
\end{equation}
where $E(s)$ denotes the embedding of sentence $s$.

\paragraph{Structural Complexity ($C_{\text{structural}}$)} Quantifies prompt decomposition into structural elements:
\begin{equation}
C_{\text{structural}}(p) = k \cdot \log\left(\frac{|p|}{k}+1\right)
\end{equation}
where $k$ is the number of distinct structural components in the prompt.

\paragraph{Task Complexity ($C_{\text{task}}$)} Evaluates the intrinsic complexity of the requested task, rated on a scale from 1 to 5:
\begin{itemize}
\item 1: Simple factual queries
\item 3: Moderate reasoning or summarization tasks
\item 5: Complex multi-step reasoning tasks
\end{itemize}

\subsubsection{Interpretation}

Lower PCM scores correspond to simpler prompts, characterized by common vocabulary and straightforward structures, while higher PCM scores indicate more intricate, challenging prompts requiring deeper comprehension or complex reasoning abilities.

\subsubsection{Customization}

The weighting coefficients $\alpha, \beta, \gamma, \delta, \epsilon$ can be adapted according to specific research contexts or task requirements, enhancing PCM's applicability and flexibility.

\newpage
\subsection{Evaluation of \textsc{PromptAnatomy's} Component Classification Accuracy}\label{sec:acc}

In this appendix, we present a comprehensive evaluation to assess the effectiveness and reliability of the proposed PromptAnatomy (PA) framework in accurately identifying and classifying semantic components within free-form prompts. This analysis serves to validate whether the PA framework can correctly dissect prompts into meaningful substructures under our defined taxonomy.

\subsubsection{Experimental Setup}

To rigorously evaluate the classification accuracy of PA, we curated a human-annotated dataset across four task domains—PubMedQA, EMEA, Leetcode, and CodeGeneration—spanning both knowledge-intensive and procedural prompts. Each prompt was manually segmented and labeled on a per-sentence basis by expert annotators according to the canonical component schema:\texttt{<Role>}, \texttt{<Directive>}, \texttt{<Additional Information>}, \texttt{<Output Formatting>}, and \texttt{<Examples>}. The annotated data served as ground truth for benchmarking model predictions.

We employed the PA dissection pipeline (Algorithm 1) using six representative LLMs:
(1) ChatGPT-4o, (2) ChatGPT-3.5, (3) Qwen2.5-14B, (4) Qwen3-8B, (5) LLaMA3.3-70B, and (6) Gemma3-12B.

\subsubsection{Metric: Classification Accuracy}

We define the  classification accuracy as: 

\begin{equation}
\text{Accuracy} = \frac{C_p}{N_p}
\end{equation}

where $C_p$ denotes the number of prompts in which all sentences are correctly classified (i.e., the predicted label set exactly matches the annotated labels for every sentence in the prompt), and $N_p$ is the total number of prompts in the evaluation set. 

A prompt is considered correct only if the model’s prediction for each sentence in the prompt exactly matches the human-annotated label set. Partial or sentence-level mismatches result in the entire prompt being marked as incorrect.

\subsubsection{Results and Analysis}
The results, summarized in Table~\ref{tab:classification_accuracy} , demonstrate strong classification performance across all models. ChatGPT-4o achieves perfect accuracy (100\%) on all four datasets, underscoring its exceptional semantic parsing capabilities. ChatGPT-3.5, Qwen2.5-14B, and LLaMA3.3-70B follow closely with average accuracies of approximately 93\%, while Qwen3-8B and Gemma3-12B achieve slightly lower scores of 92\% and 88\%, respectively.

\begin{table}[ht]
\centering
\small
\caption{Classification Accuracy (\%) across different datasets and models}
\label{tab:classification_accuracy}
\begin{tabular}{lccccc}
\toprule
\textbf{Model} & \textbf{PubMedQA} & \textbf{EMEA} & \textbf{Leetcode} & \textbf{CodeGeneration} & \textbf{AVG} \\
\midrule
ChatGPT-4o          & 100.00 & 100.00 & 100.00 & 100.00 & 100.00 \\
ChatGPT-3.5         & 89.00  & 95.00  & 96.00  & 93.00  & 93.00  \\
Qwen3-8B            & 91.00  & 94.00  & 93.00  & 89.00  & 92.00  \\
Qwen2.5-14B         & 88.00  & 94.00  & 95.00  & 93.00  & 93.00  \\
LLaMA3.3-70B        & 91.00  & 97.00  & 95.50 & 90.30 & 93.45 \\
Gemma3-12B          & 86.00  & 89.00  & 94.50 & 82.50 & 88.00  \\
\bottomrule
\end{tabular}
\end{table}

These results confirm that the PA framework generalizes well across model scales and architectures, achieving consistent accuracy above 85\% even in smaller models. The use of local contextual windows appears to significantly improve the model's ability to disambiguate component roles, especially for structurally ambiguous prompts.

Notably, all models perform robustly across datasets with differing domain styles, indicating that prompt structure is learnable and recognizable irrespective of topical content.

\newpage
\subsection{Visualization of Table \ref{tab:asr_components}}\label{sec:heatmap}

To facilitate a more intuitive understanding of the heterogeneous adversarial robustness across prompt components and perturbation types, we present heatmap visualizations of Table~\ref{tab:asr_components} for each evaluated LLMs. Each heatmap illustrates the attack success rate (ASR) under different component-wise perturbations, grouped by model and dataset.

\begin{figure}[htbp]
    \centering
    \includegraphics[width=0.9\linewidth]{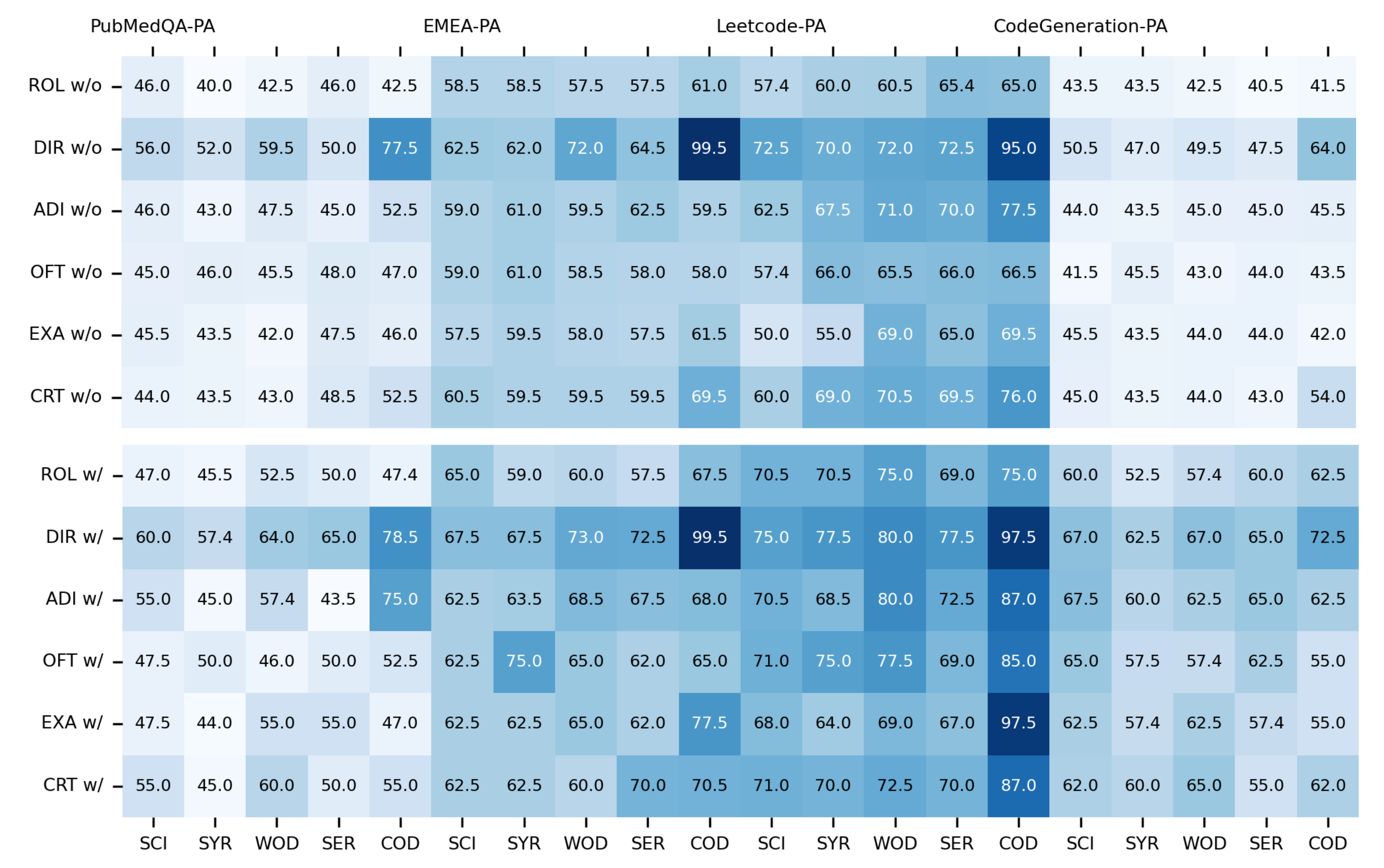}
    \caption{GPT-4o}
    \label{fig:heatmap-gpt}
\end{figure}

\begin{figure}[htbp]
    \centering
    \includegraphics[width=0.9\linewidth]{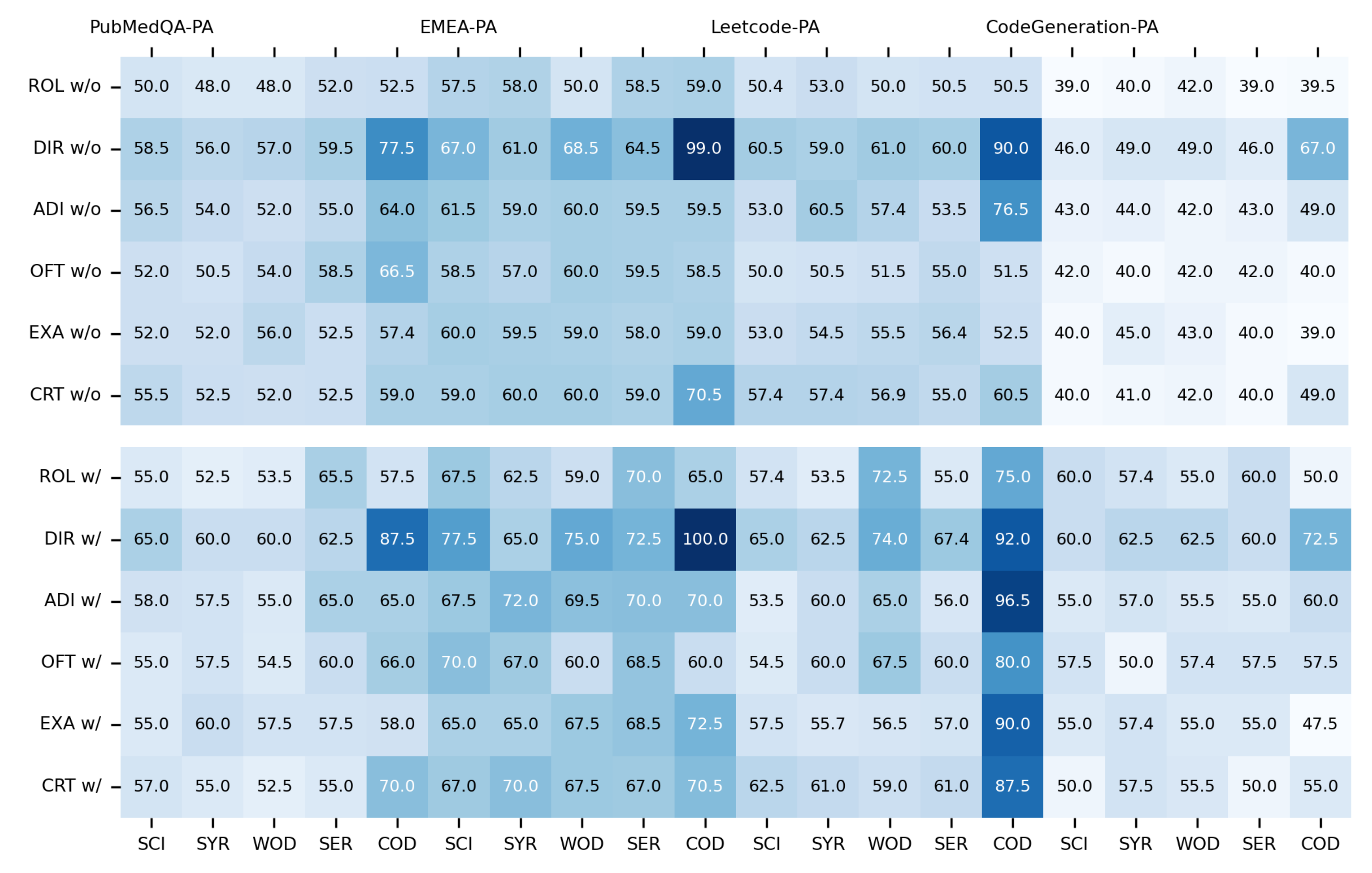}
    \caption{Claude3.7}
    \label{fig:heatmap-claude}
\end{figure}

\begin{figure}[htbp]
    \centering
    \includegraphics[width=0.9\linewidth]{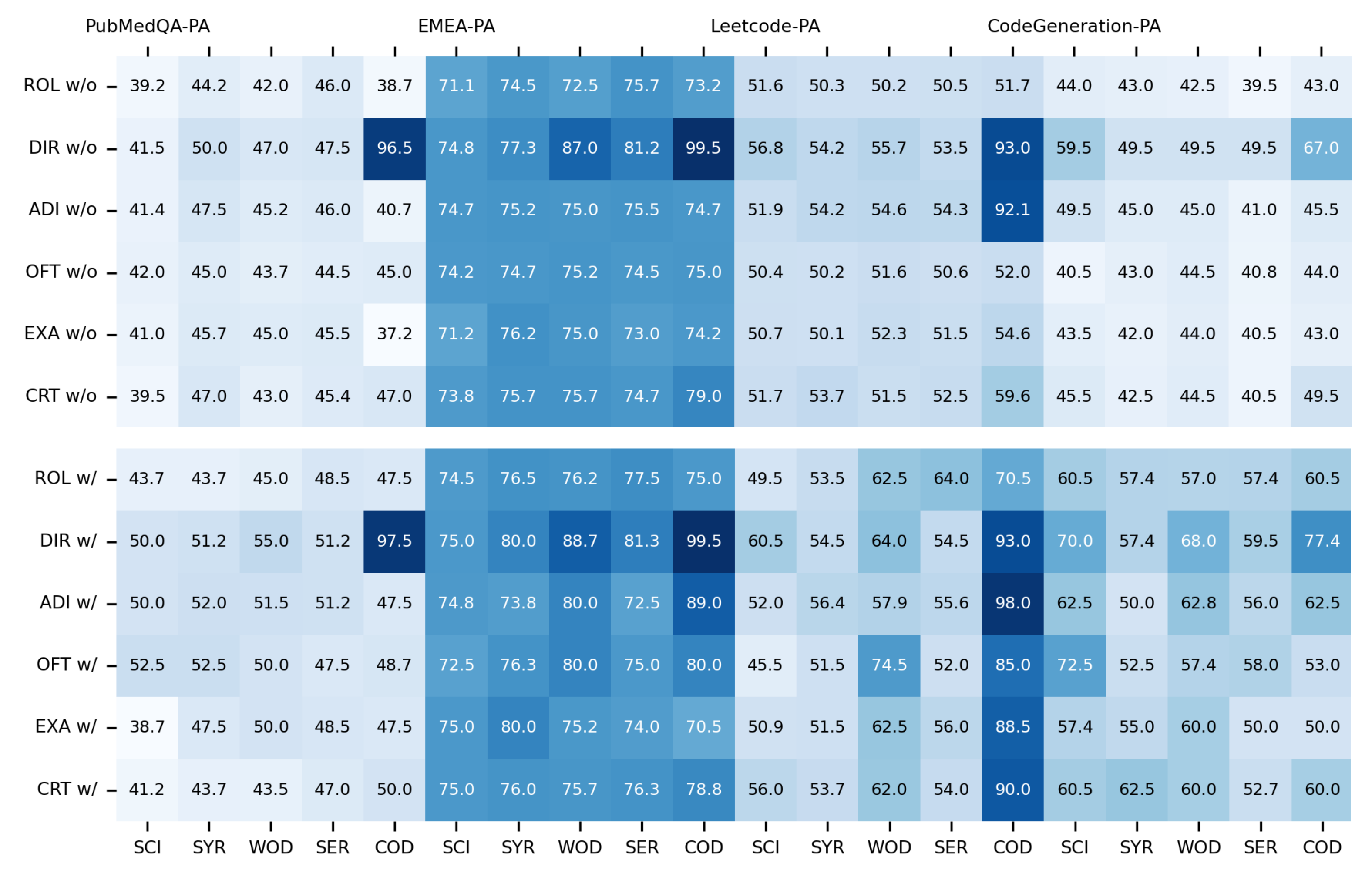}
    \caption{Qwen2.5}
    \label{fig:heatmap-qwen}
\end{figure}

\begin{figure}[htbp]
    \centering
    \includegraphics[width=0.9\linewidth]{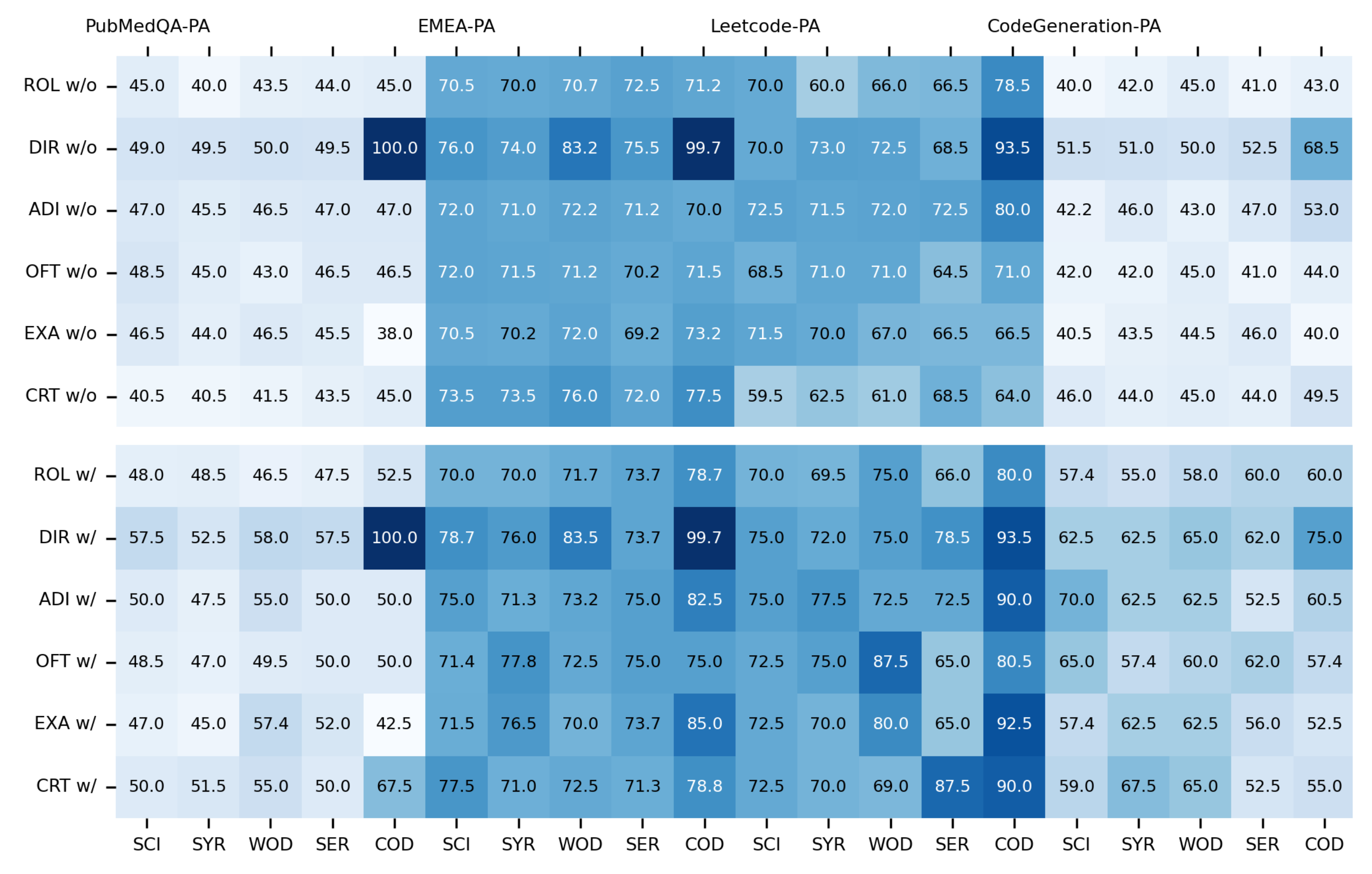}
    \caption{LLaMA3-70B}
    \label{fig:heatmap-llama}
\end{figure}

\begin{figure}[htbp]
    \centering
    \includegraphics[width=0.9\linewidth]{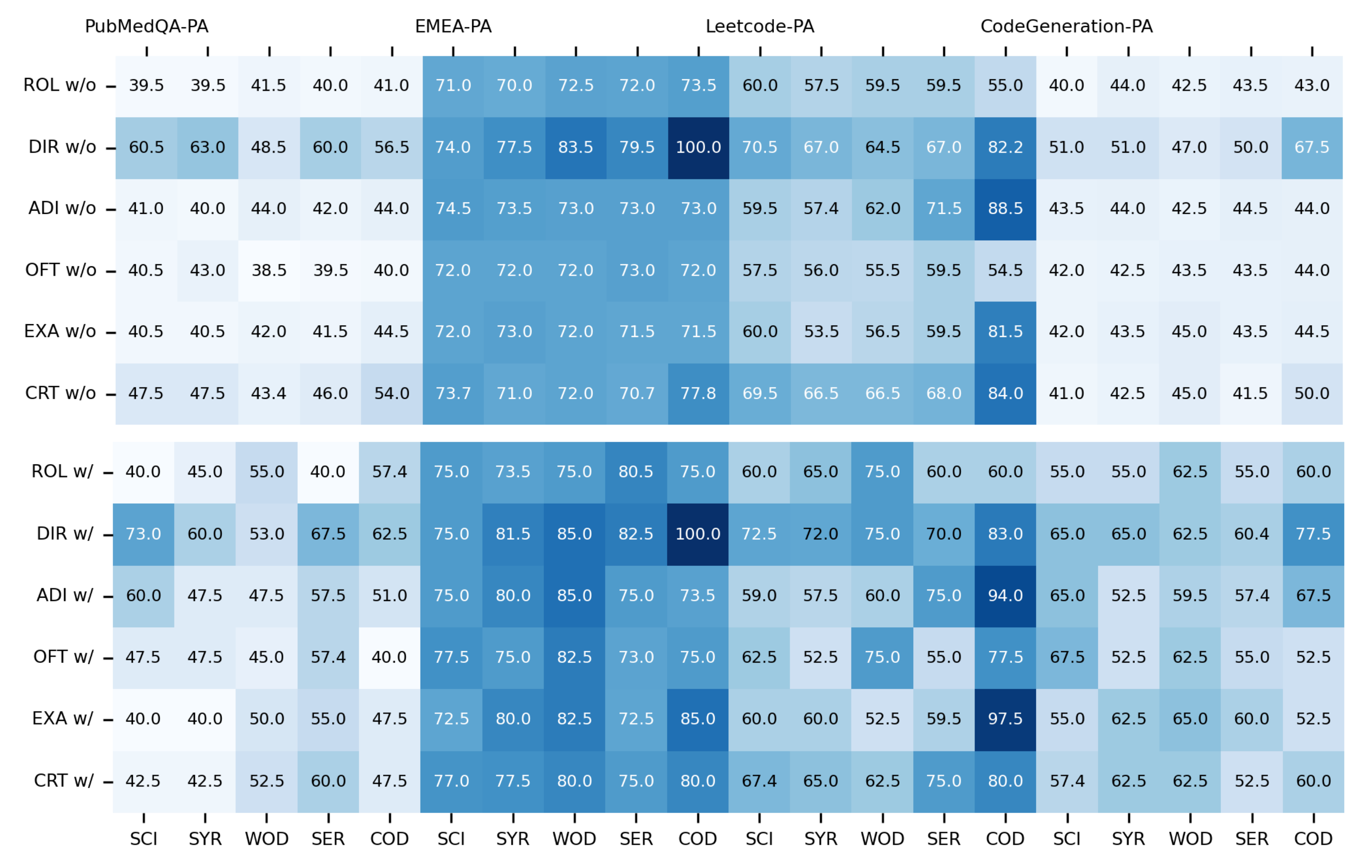}
    \caption{Gemma3-12B}
    \label{fig:heatmap-gemma}
\end{figure}

\end{document}